\documentclass{article} 
\usepackage{iclr2027_conference,times}

\usepackage{amsmath,amsfonts,bm}

\def\eqref#1{equation~\ref{#1}}

\def\1{\bm{1}}

\DeclareMathAlphabet{\mathsfit}{\encodingdefault}{\sfdefault}{m}{sl}
\SetMathAlphabet{\mathsfit}{bold}{\encodingdefault}{\sfdefault}{bx}{n}

\usepackage{amsfonts}

\usepackage{amsmath, amssymb, amsthm}

\theoremstyle{plain}

\theoremstyle{definition}

\theoremstyle{remark}

\usepackage{hyperref}
\usepackage{url}
\usepackage{tikz}
\usepackage{longtable}
\usepackage[normalem]{ulem}

\usepackage[T1]{fontenc}
\usepackage{xcolor}
\usepackage{colortbl}

\usepackage{xcolor}
\usepackage{listings}
\usepackage[most]{tcolorbox}
\usepackage{booktabs}

\definecolor{promptbg}{HTML}{F7F7F8}
\definecolor{promptframe}{HTML}{D0D0D0}
\definecolor{prompttitle}{HTML}{F0F0F2}

\lstdefinestyle{promptstyle}{
  basicstyle=\ttfamily\footnotesize,
  columns=fullflexible,
  keepspaces=true,
  breaklines=true,
  breakatwhitespace=false,
  showstringspaces=false,
  upquote=true,
  tabsize=2
}

\newtcblisting{promptbox}[2][]{
  enhanced,
  breakable,
  colback=promptbg,
  colframe=promptframe,
  coltitle=black,
  colbacktitle=prompttitle,
  fonttitle=\bfseries,
  title={#2},
  listing only,
  listing options={style=promptstyle},
  arc=2mm,
  boxrule=0.5pt,
  left=1mm,
  right=1mm,
  top=1mm,
  bottom=1mm,
  #1
}

\title{Beyond Selected Breakthroughs: Evaluating Language Models on Randomly Sampled Mathematical Conjectures}

\author{%
Helen Jenne$^{1}$\thanks{Corresponding author: \texttt{helen.jenne@pnnl.gov}}
\quad
  Davis Brown$^{3}$
  \quad
  Jesse He$^{1,2}$
  \quad
  Henry Kvinge$^{1,4}$
  \\
  \\
  $^1$ Pacific Northwest National Laboratory, $^2$ University of California San Diego\\
  $^3$ University of Pennsylvania, $^4$ University of Washington
}

\iclrfinalcopy 
\begin{document}

\maketitle

\begin{abstract} 
The last year has seen a shift in AI-assisted mathematics, with frontier language models producing proofs of and counterexamples to long-standing open conjectures. However, studying only the current set of major breakthroughs provides an incomplete picture of LLM capabilities and the computational resources needed to achieve reported results. To address this, we introduce a new dataset, OpenConjecture, which collects conjectures from mathematics papers published on the arXiv since February 2026. We use these conjectures as a source of open mathematical problems, and apply frontier GPT models to a random sample of 200 of these conjectures over three passes using a reproducible protocol without problem-specific mathematical guidance. Rather than attempt expert verification of these candidate solutions, we analyze the types of problems for which LLMs produce solutions and the characteristics of these solutions. The models we test produce candidate solutions for roughly 60\% of the conjectures, with counterexamples about 1.3 times more common than proofs. We estimate solution novelty, assess how solutions cite and relate to existing literature, and analyze whether the proposed resolution relies on either an error in or underspecification of the conjecture. We find that models sometimes propose resolutions that are outside the authors' intent (e.g., proposing the trivial group as a counterexample) and that a number of proofs follow closely from existing results, but that some resolutions exhibit at least moderate novelty. Our findings underscore the need for evaluations that go beyond correctness to assess fidelity to the intended statement, attribution, and the nature of the mathematical contribution.
\end{abstract}

\section{Introduction}

The last year has seen a dramatic shift in the capabilities of frontier AI systems in research-level mathematics. The disproof of the Erd\H{o}s unit distance conjecture \citep{openai2026discretegeometry}, Jacobian conjecture \citep{sircar2026mathematicians}, the recent Navier–Stokes blowup result \citep{openai2026navierstokes}, and other announcements of AI-assisted advancements \citep{openai2026tenadvancements} have demonstrated that AI systems are able to produce proofs, as well as counterexamples, to difficult mathematical problems. This has become clear not only from announcements from frontier labs
but also through carefully curated benchmarks such as First Proof \citep{abouzaid2026first, abouzaid2026proofsecondbatch} and Benchmarks in Leipzig \citep{balakin2026benchmarks}. These and similar benchmarks have helped assess model performance on collections of problems whose solutions are known but not publicly available online, and often provide detailed information about prompting and cost. 

Still, the rapid progress has left many questions unanswered. Do LLMs more often disprove conjectures by finding counterexamples, or prove them? How many conjectures can be resolved with a single na\"{\i}ve prompt, and how often is repeated prompting or domain expertise needed?  Are LLM solutions primarily driven by broader awareness of the literature such that little novelty is typically introduced? We are interested in how these and related questions are answered on a {\em random} sample of conjectures.

This motivates OpenConjecture, a dataset of conjectures from recent arXiv mathematics preprints. In this paper we introduce this dataset and use it as a source of conjectures drawn `from the wild'. We study 200 conjectures, one drawn from each of 200 randomly sampled papers. We then prompt several versions of GPT models, specifically GPT-5.6 Sol, GPT-5.6 Sol Pro, and GPT-6 Astra Pro, to produce proofs or counterexamples across three passes, each of which receives the previous pass's unresolved attempts as context. We analyze the resulting output to better understand the qualities of LLM mathematical ability beyond correctness. As such our purpose is not to announce a long list of new results, but rather see what we can learn about the types of solutions these systems produce. Out of courtesy to the reader, we keep the LLM-generated mathematics to a minimum, including examples only as needed to illustrate our main points.

We find that the models' proposed conjecture resolutions vary significantly in what they contribute mathematically. For example, of the proposed counterexamples from the first and second attempts that passed an LLM verification review, roughly 60\% were classified as ``off-the-shelf'' and the remainder as ``standard constructions''. Some of the counterexamples appeared to take advantage of missing hypotheses in the conjecture statement rather than offer a mathematically interesting construction. In contrast, the counterexamples from the third pass typically involved nontrivial modifications or more involved arguments for why the proposed object does in fact refute the conjecture. Findings like these illustrate that technical correctness is just one dimension of LLM mathematical output.

Concretely, this paper makes the following contributions: 
\begin{itemize}
\item We introduce the OpenConjecture dataset, which gathers conjectures from recent arXiv preprints and makes them publicly available for analysis. 
\item We prompt frontier GPT models, in three passes, to solve a sample of 200 conjectures drawn from this dataset. We find that models propose counterexamples to 70 conjectures and proofs to 55 conjectures, and that 84\% of these proposed resolutions hold up to LLM review. 
\item We analyze the mathematical content of proposed solutions, including their length (both in terms of word count and number of definitions, lemmas, theorems, etc.), the novelty of counterexample constructions and their alignment with author intent, and the relationship of proposed proofs to the existing literature.
\end{itemize}

While we do not attempt expert verification or formalization of the conjecture attempts or use this as a benchmark to compare models, we encourage extensions of this work: OpenConjecture is available on HuggingFace\footnote{\url{https://huggingface.co/datasets/OpenConjecture/openconjecture}.
} and model responses to the 200 sampled conjectures will be made available separately.

\section{Related Work}

The field of AI for math has grown significantly in recent years. We refer the reader to \citep{ju2026ai} for a survey and \citep{leeAwesomeAIMath} for a curated list of papers in the area. Here, we focus on the related work that most directly informs our methodology. 

Similar to OpenConjecture, {\bf LemmaBench} \citep{peyronnet2026lemmabench} and {\bf arXivMath} \citep{dekoninck2026arxivmath} also source problems from recent arXiv papers. Both cite the high risk of data contamination as a motivator. LemmaBench rewrites lemmas from recent arXiv preprints into self-contained statements. 
arXivMath was generated by asking an LLM to create questions from arXiv abstracts, rejecting abstracts that do not give rise to a self-contained problem. Then they have several additional filters and preprocessing steps: they use an LLM to add missing conditions, and to check (from the full paper in markdown) whether the answer is inferrable from prior work. Like these datasets, OpenConjecture and our corresponding analysis aims to capture mathematics `in the wild'. Unlike these datasets, however, OpenConjecture collects conjectures made by human mathematicians, rather than lemmas or questions synthetically generated from an abstract.

The {\bf First Proof} batches each consist of 10 research-level questions. The questions span a variety of fields in mathematics and are problems that arose in the research process of the contributors but whose proofs had not been published. The authors describe the first batch of 10 problems \citep{abouzaid2026first} as an informal experiment, and the second batch of 10 problems as a more formal benchmark. In the second batch, they tested and graded four systems: Chat GPT-5.5 Pro and three harnesses.
Solutions were refereed and graded by experts. Some particularly noteworthy findings included: (i) at least seven of the 10 problems received at least one passing grade; (ii) in three problems, the argument was not the same as the known solution; (iii) in several solutions, attribution failures were significant enough that if a human submitted the solutions it would be considered plagiarism. 

Other notable research-level benchmarks include Benchmarks in Leipzig \citep{balakin2026benchmarks}, which consists of 100 research-level questions with unique, well-defined answers, contributed from 49 mathematicians, IMProofBench \citep{schmitt2025improofbench}, Soohak \citep{son2026soohak}, Riemann-Bench \citep{garre2026riemann}, and FrontierMath \citep{glazer2024frontiermath}.

\section{OpenConjecture}
\label{sect:open-conjecture}

The OpenConjecture dataset is created by ingesting recent arXiv math preprints and extracting conjecture blocks. We use GPT-5 Mini to label if the conjecture is actually being posed by the authors, as opposed to a conjecture that is being stated because the paper has resolved it or given as background motivation. We also use GPT-5 Mini to estimate the mathematical interestingness and near-term solution viability of conjectures. These statistics may be of independent interest because of what they can tell us about the current landscape of mathematical conjectures.
The first OpenConjecture snapshot was published March 17, 2026, with most papers uploaded after February 2, 2026. It is updated weekly. Code to generate snapshots can be found at \url{https://github.com/davisrbr/conjectures-arxiv}.

\begin{figure}
\centering
\includegraphics[width=.6\textwidth]{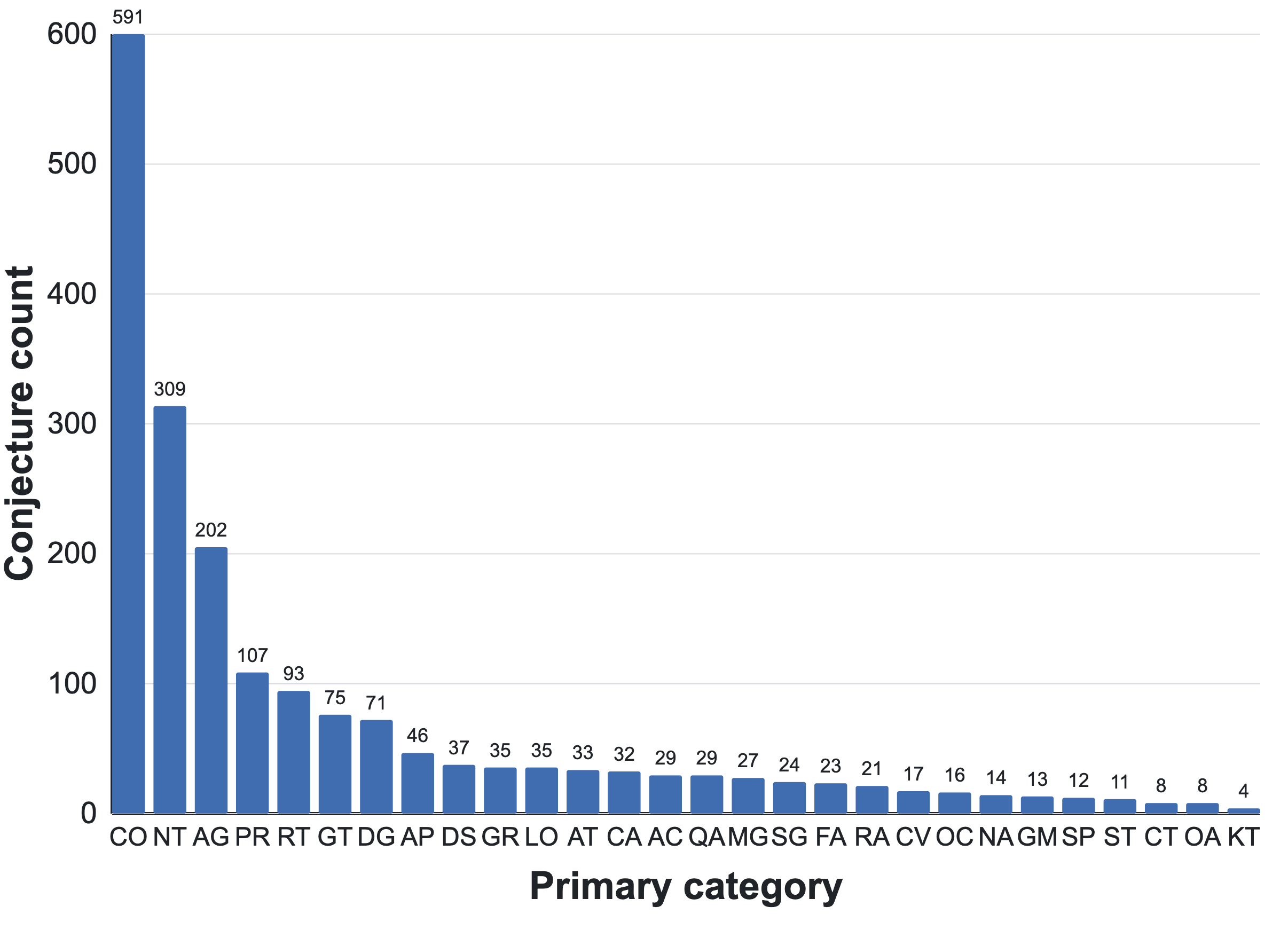}
\caption{All open conjectures with a math primary subject classification in the June 21st snapshot of OpenConjecture. Categories with a single conjecture are omitted.}
\label{fig:conjecture-category}
\end{figure}

For our case study, we use the June 21st snapshot, which contains 3030 conjectures (from 1675 papers) that were labeled as open by GPT-5 Mini. We filtered an additional 769 conjectures that were from papers that did not have a math primary subject classification. These papers were cross-listings or updates rather than papers from the intended window.

In arXivMath \citep{dekoninck2026arxivmath} they found that even after their initial filtering, a review of 20 questions showed that 30\% were potentially answerable using prior work cited. This motivated us to apply a second round of open/not-open filtering on the 2,261 downselected conjectures using GPT-5.4. In this step, 337 conjectures were reclassified as not real, leaving us with 1,924 conjectures labeled as open. 

The majority of the remaining 1,924 conjectures come from combinatorics, number theory, and algebraic geometry, as shown in Figure~\ref{fig:conjecture-category}. The overall population of papers by subfield in the entire snapshot is shown in Figure~\ref{fig:paper_primary_category} in Appendix~\ref{appendix:open_conjecture}. While combinatorics is the most prevalent, there are four categories with more papers than number theory and algebraic geometry (analysis of PDEs, optimization and control, numerical analysis, and probability). Other descriptive statistics related to the dataset and further preprocessing details can be found in Appendix~\ref{appendix:open_conjecture}.

\section{Finding Candidate Solutions to Open Conjectures}

In this section we describe our methodology for sampling conjectures and prompting the models to solve the conjectures. 

\paragraph{Sampling} We sampled 200 papers uniformly from the papers containing at least one conjecture after the filtering process described in Section \ref{sect:open-conjecture}. We then sampled one conjecture uniformly from each paper. This avoided the scenario where some conjecture-heavy works are overrepresented (see Table \ref{table:per-paper-conjecture-count} in Appendix~\ref{appendix:additional_plots}). We chose this approach, rather than sampling a fixed number of conjectures by primary category, because we wanted our results to reflect the global population of recent arXiv math preprints containing conjectures, without overweighting subfields that publish few conjectures.

\paragraph{Solver} We used up to three sequential solver passes: GPT-5.6 Sol in the first pass, GPT-5.6 Sol Pro in the second, and GPT-6 Astra Pro in the third. Only conjectures that received a verdict of ``unresolved'' by the solver proceeded to the next pass. All runs used reasoning effort \texttt{xhigh}. This design was intended to approximate a workflow of a mathematician using frontier models under cost constraints: starting with a strong but not the most expensive model, and increasing the strength of the model in subsequent attempts if the conjecture remains unresolved. 

In the first pass we provided the conjecture, the full paper, and the instruction to ``make a sustained attempt to prove or refute'' the conjecture to GPT-5.6 Sol. The model could return one of three verdicts: \texttt{proof}, \texttt{counterexample}, or \texttt{unresolved}. It was also instructed to cite any external results used. The complete prompt is given in Appendix~\ref{appendix:prompts}. We kept the prompt relatively simple, following the example of FirstProof \citep{abouzaid2026first}, to avoid imposing structure on the LLM outputs which could divert effort from attempting the problem. 

If the conjecture was marked unresolved after the first pass, GPT-5.6 Sol Pro was prompted to resolve the conjecture. To do this, it was given the conjecture with necessary context (consisting of definitions, notation, and relevant statements required to understand the conjecture), along with the exact output of the first pass, and instructions to produce a proof or counterexample. If the conjecture was still unresolved after the second pass, we prompted GPT-6 Astra Pro to resolve the conjecture using the same input as the second pass, along with the second pass response. This pass was motivated by our initial review of second pass attempts, in which it seemed the models made nontrivial progress even though they did not resolve the conjecture. The third pass examined whether a stronger model, given the preceding attempts, would produce a candidate proof or counterexample. 
Web search was permitted in the first pass, but we removed it for the second and third pass after initial review of first pass responses showed examples that seemed more focused on literature review than thinking about the problem. 



\section{Results and Analysis}

\begin{figure}
\centering
\includegraphics[width=0.35\textwidth]{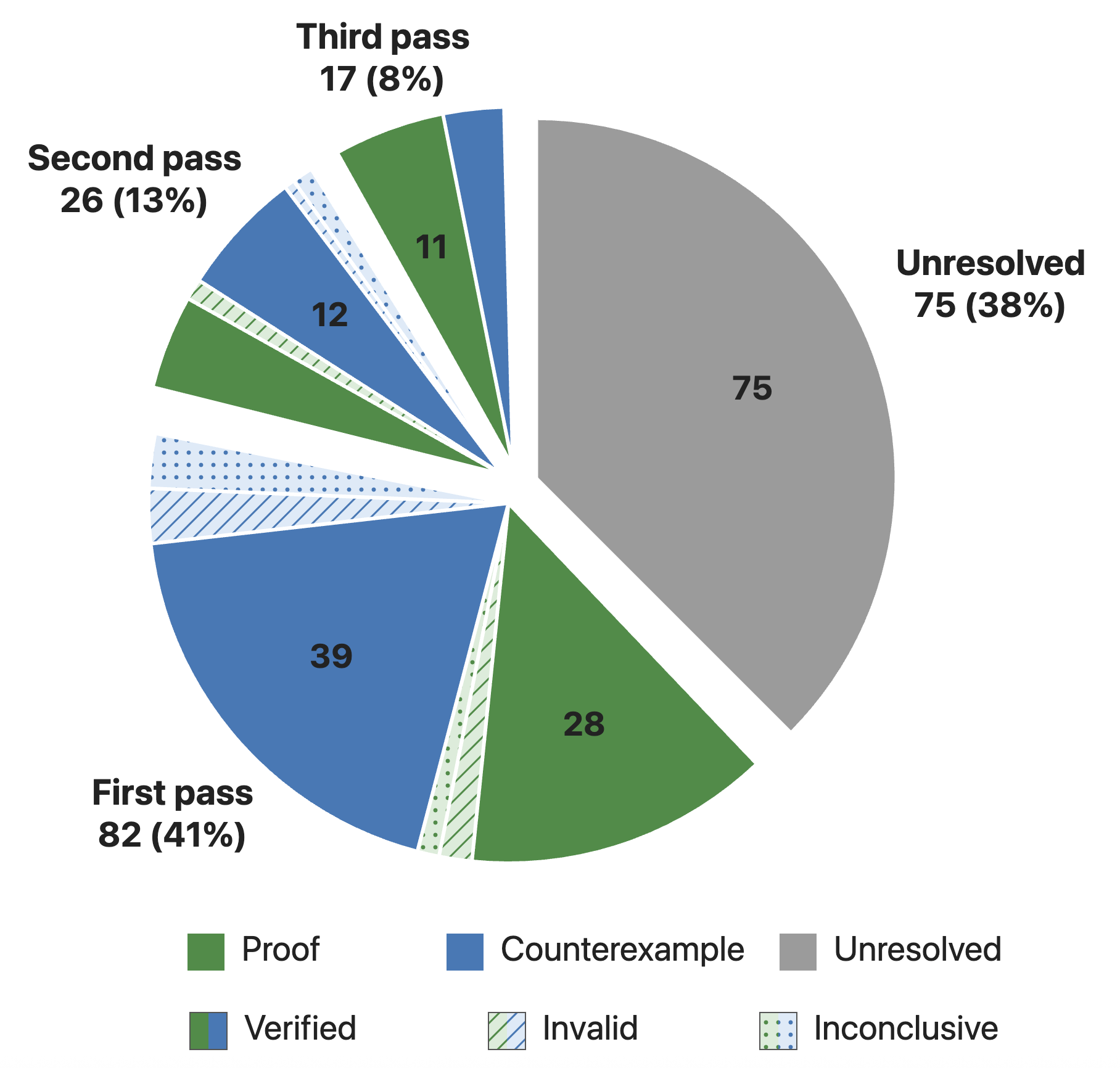}
\includegraphics[width=0.6\textwidth]{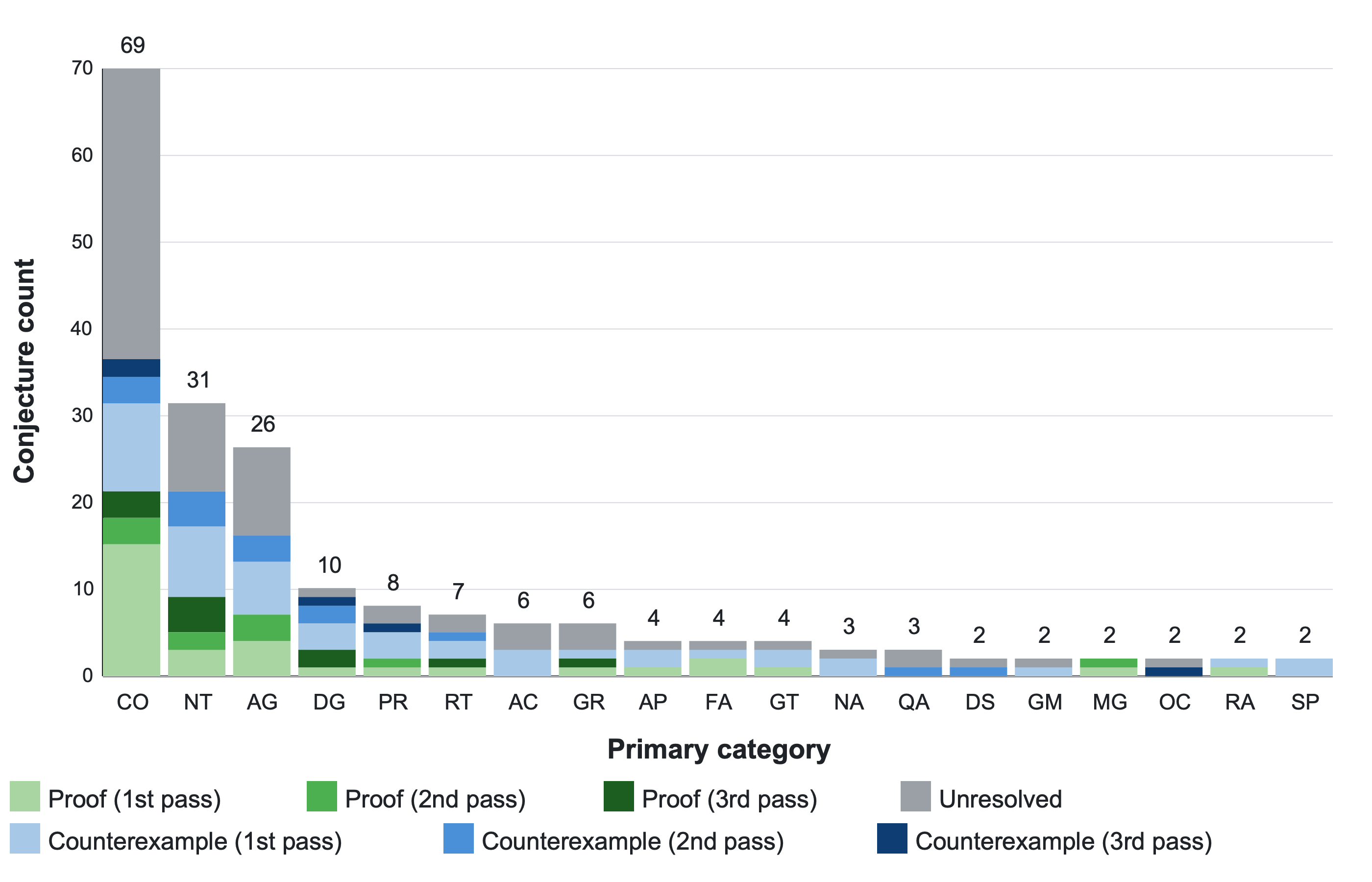}
\caption{Left: Final status of conjectures and the pass in which a resolution was proposed; hatching indicates the LLM verification outcome.
Right: Final status by math subject classification, with verification status omitted. The seven single-conjecture categories are not shown (AT, CT, CV, HO, LO, OA, and SG, with outcomes: three counterexamples, two proofs, and two unresolved). }
\label{fig:overall}
\end{figure}

In total, the three passes produced 70 counterexamples and 55 proofs to the 200 conjectures. Not all of these stood up to LLM-based verification. Figure~\ref{fig:overall} shows the overall results, both by math subject classification and by the pass in which a resolution was proposed. Among the counterexamples, 49 were found in the first pass, 15 in the second, and 6 in the third. Among the proofs, 33 were found in the first pass, 11 in the second, and 11 in the third. The number of proposed counterexamples outnumber proofs approximately 60/40 in the first two passes, but by the third pass, the number of proposed proofs outnumbered counterexamples. In this section, we give an overview of results, discussing the LLM correctness review, our findings about the quantity and nature of citations in proposed resolutions, and the length and cost of responses, before discussing the counterexamples and proofs in more detail. 

\paragraph{LLM review for correctness}  For the proposed resolutions, we asked a new instance of the same model to verify the correctness of the attempt. The reviewer model was given the proposed proof or counterexample and the full paper. The prompt used can be found in Appendix~\ref{appendix:prompts}. All third pass proofs and counterexamples, and between 80 and 85\% of first and second pass proofs and counterexamples, passed this check.  
We manually reviewed the verification to check for obvious issues (such as a verifier run refuting a proof with an edge-case counterexample that was likely misaligned with the intention of the authors), but true verification would require expert review and is outside of scope for our study. Our analysis of proofs and counterexamples in the sections below focuses on those that passed the validation check. 

\paragraph{Resolution citations and other references}

In FirstProof Second Batch~\citep{abouzaid2026proofsecondbatch}, the authors noticed models often failed to appropriately attribute results. This motivated our decision to explicitly request citations in the first pass prompts. This was effective at getting models to include citations. Under a broad definition of a citation as an explicit reference to an external work, theorem, named result, etc, we found that the first pass responses included an average of 3.3 citations and a median of 2 distinct citations. (This count does not include the arXiv preprint which contained the conjecture, which was cited in 93\% of responses.) Responses that were unresolved had the most citations; counterexamples were most likely to have no citations other than the source paper.  These citations were most often arXiv preprints (29\%) or journal articles or other papers (24\%), but also included named results (20\%) and constructions, formulas, or similar (11\%) (Figure~\ref{fig:first_pass_citations} in Appendix~\ref{appendix:additional_plots}). 

The second and third pass responses also included citations under this broad definition (both had a median of 2 distinct citations, with means of 2.8 and 2.2, respectively) but the type of citations were very different compared to the first pass responses: about 60\% were named theorem or results, and the remainder were different sub-categories of constructions, formulas, or other criteria. 
The prompt the second- and third-pass solvers received included instructions not to use web search or discuss the literature status of the conjecture, and it is possible that (i) the distinction between giving proper attribution and discussing the literature status is subtle and (ii) not being able to use web search and check the accuracy of a citation makes a model less likely to include a citation.

Requesting citations in the first pass came with a trade-off: responses seemed to be more focused on literature review/search and discussion of the status of the conjecture rather than working on the problem.
We hypothesized that specifically requesting precise citations and providing web search exacerbated this issue. Indeed, the models used web search more than we expected: in the first pass, the median number of web search calls per conjecture was 18 and the median number of web search queries was 40. The distribution of web search calls is shown in Figure~\ref{fig:web_search_calls} in Appendix~\ref{appendix:additional_plots}. Accordingly, the median number of input tokens in this pass was 131,783, as opposed to 6,309 in the second pass (recall also that the first pass was the only pass to receive the full paper as opposed to just the relevant conjecture context).

\paragraph{Response length} We measured the length of responses in two ways: by the prose word count (the number of words in the natural-language portions of the final response, excluding mathematics enclosed in standard LaTeX delimiters, numeric citation markers, and arXiv identifiers), and the total number of claims with proof. The third-pass responses are the longest by both measures. Unresolved responses tend to be longer in word count compared to proofs and counterexamples but less so in the second pass (Figure~\ref{fig:length} in Appendix~\ref{appendix:additional_plots}). The prose word count is shown by subfield in Figure~\ref{fig:length_subfield} in Appendix~\ref{appendix:additional_plots}; there seems to be more variability in response length within the subfields than between subfields. In contrast to the number of lemmas, theorems, propositions, and corollaries, responses contained very few definitions that were introduced by the solver (as opposed to definitions from the source paper, external literature, or standard mathematical background). 
This is consistent with claims that models tend to infrequently produce solutions and counterexamples that are novel, although the introduction of definitions is only one possible indicator of novelty. 

\paragraph{Cost} Plots showing input and output token usage for the first, second, and third passes are shown in Appendix~\ref{appendix:additional_plots}. The median number of input tokens/output tokens was 131,783/24,956, 6,309/34,522, and 9,727/29,022 for the first, second, and third passes, respectively, so the median cost by pass was approximately \$0.97, \$0.41, \$1.71 at the time of writing. 





\subsection{Counterexamples}

\begin{figure}
\centering
\includegraphics[width=.45\textwidth]{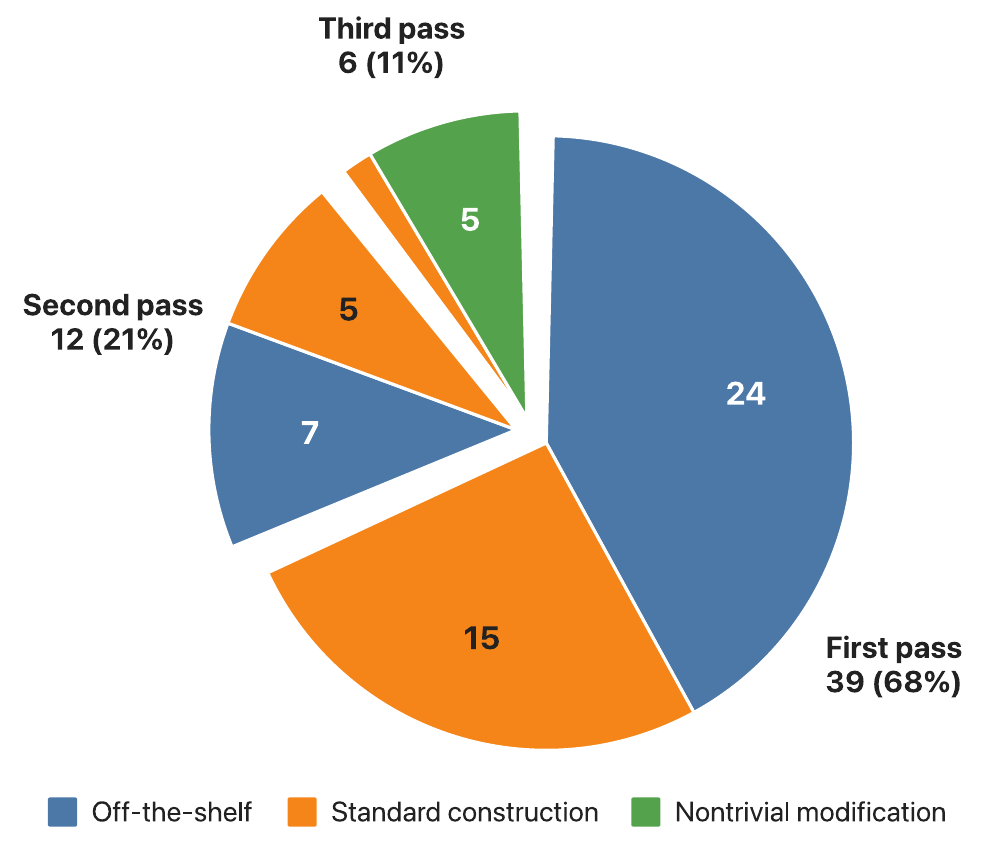}
\includegraphics[width=0.45\textwidth]{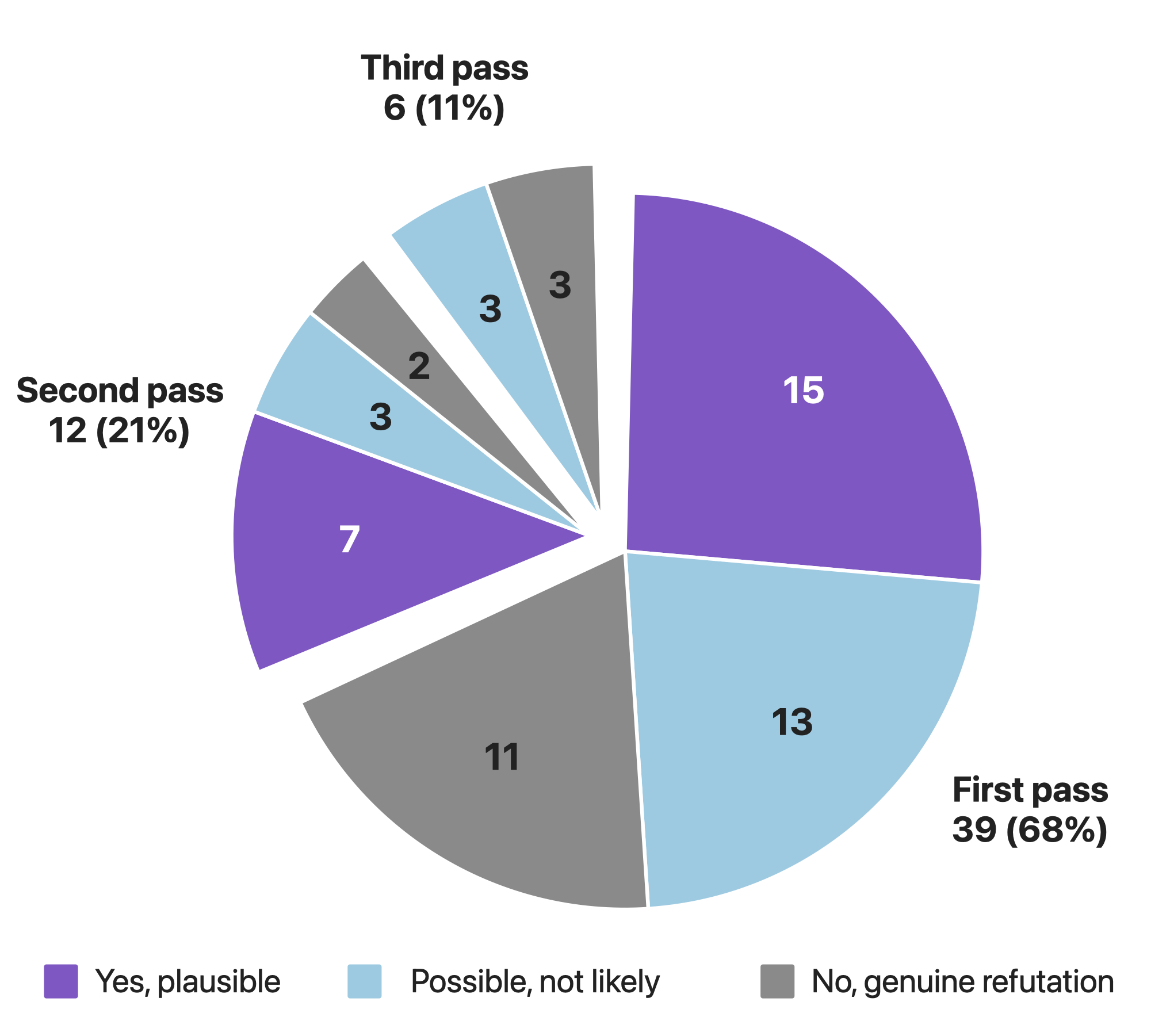}
\caption{Left: LLM classification of object novelty for counterexamples that were `verified` by the review pass. Counterexamples are classified as `off-the-shelf', `standard construction', or `nontrivial modification'. Right: LLM responses to the question `is it plausible that the conjecture's authors implicitly excluded this case' (i.e., the conjecture is misstated rather than genuinely refuted)?}
\label{fig:cex}
\end{figure}

Our review of counterexamples is focused around two questions:
\begin{itemize}
\item[(i)] How often was a conjecture refuted by an ``off-the-shelf'' counterexample, meaning an example a specialist in the field would have on their standard list of things to try?
\item[(ii)] How often do counterexamples rely on errors or underspecification in conjecture statements?
\end{itemize}

The motivation for the first question comes from the fact that in many fields of mathematics, there are ``standard'' counterexamples documented in books like Counterexamples in Topology \citep{steen1978counterexamples} and Counterexamples in Analysis \citep{gelbaum2003counterexamples}. We were also influenced by a recent blog post by Timothy Gowers discussing the kinds of mathematics problems LLMs are good at, which listed different methods for counterexample search \citep{gowers2026maths}. Furthermore, in our initial manual review of counterexamples, we noticed many straightforward counterexamples in the first pass responses. 

This observation also motivated question (ii) since the simplicity of some counterexamples was so extreme that it seems likely that LLM interpretation of the problem must have been misaligned with the intent of the authors. For example, counterexamples to some of the conjectures in combinatorics included the cycle graph on five vertices and the complete graph on five vertices. Another example, given to a number theory conjecture, was the trivial group. The model responses themselves also seemed to suggest that the provided counterexample may have been implicitly excluded by the authors. Specifically, we noticed counterexample explanations sometimes included a paragraph titled ``Scope'', with language similar to the following: ``The counterexample refutes the conjecture exactly as stated.'' These types of responses frequently went on to list amendments to the conjecture that would make the counterexample no longer hold. We hypothesize that at least in some cases this amendment was what was intended by the paper authors. In another example, the model labeled its counterexample ``literal counterexample'': a single-vertex graph.

\subsubsection{Methods} 
After a manual review of counterexamples, we used an LLM reviewer to produce the results shown in Figure~\ref{fig:cex}. For each counterexample, we prompted a new instance of the same model that produced the counterexample to classify the object as {\em off-the-shelf} (an example a specialist would have on their standard list of things to try), a {\em standard construction} (obtainable from off-the-shelf examples by standard modifications within the subfield), a {\em nontrivial modification} (built from known ingredients but requiring a non-obvious choice), or {\em genuinely novel} (introducing an object or construction not reducible to known ingredients). We also asked the model to give an assessment of the {\em novelty of the argument} supporting the counterexample. This aimed to assess the difficulty of the argument needed to justify the counterexample. The review prompt also asked the reviewing model to assess whether it is plausible that the authors of the conjecture implicitly excluded this counterexample and/or the conjecture was too imprecise. 


The complete prompt, which included several other questions including the overall novelty, is given in Appendix~\ref{appendix:reviewer_prompts}.

\subsubsection{Results}

\paragraph{Novelty of counterexamples} Figure~\ref{fig:cex} (left), which show LLM judgements of novelty for the verified counterexamples, agree with informal observations we made during manual review. Both the first pass and second pass counterexamples are judged to be about 60\% ``off-the-shelf'', and the remainder are classified as ``standard construction.'' In contrast, most third pass counterexamples were given ratings of ``nontrivial modification'' (5/6).  
The third pass counterexample responses were also longer than counterexample responses from the first two passes, in both word count and the combined total number of claims proved
(Figure~\ref{fig:length} in Appendix~\ref{appendix:additional_plots}), and these verifications were more often assessed as nonroutine by LLM reviewers. 


\paragraph{Counterexamples exploiting missing assumptions or mistakes} Figure~\ref{fig:cex} (right) shows the fraction of counterexamples that a reviewer judges to be the result of imprecision in the conjecture formulation. This makes up a significant portion of first-pass and second-pass counterexamples. Some excerpts from model reviews which are labeled as ``very plausibly'' due to imprecision include:
\begin{itemize}
\item ``This appears to expose a convention mismatch rather than defeat the intended representation-theoretic conjecture.''
\item ``The counterexample therefore diagnoses a misstatement or order-of-operations error rather than overturning the intended linearized double-shuffle conjecture.''
\item ``Consequently, this looks more like exploitation of a missing projectivity/properness hypothesis than a counterexample to the intended geometric conjecture.'' 
\end{itemize}

{\bf In summary,} the results suggest that while some counterexamples are off-the-shelf or the result of imprecision, some likely represent genuine novelty.
Reviewing the counterexamples was also instructive about datasets of conjectures ``from the wild''. The number of conjectures that could be refuted in a single pass with a counterexample classified as a standard construction or off-the-shelf may reflect that open conjectures presented in papers are not exclusively problems that have been attempted and failed, but also include problems that have not been attempted, and problems where the author(s) expect that progress can be made \citep{cohn2026technicaldebt}.
That counterexamples often seem to rely on errors or underspecification may be at least partially explained by the fact that many mathematical conjectures are not self-contained: they rely on context stated elsewhere in the paper or possibly on implicit assumptions. 
Consequently, extracting the relevant context needed for a conjecture requires some work, and even then oversight may be necessary to ensure that the LLM does not produce uninteresting counterexamples that violate the spirit of the conjecture.
 These findings are relevant to researchers making benchmarks using problems from arXiv papers and using such benchmarks to assess model capability.


\subsection{Proof Analysis}

In this section we focus on two aspects of the proofs and unresolved responses: their relationship to existing results, and the extent to which later attempts used the results and strategies developed in earlier passes.

\paragraph{Relationship to existing literature} In our initial review of proofs from the first pass, we found that a surprising number of conjectures had been resolved in a revision of the source preprint or a subsequent preprint by different authors\footnote{These cases illustrate the limitations of the literature-status filtering used in the dataset construction.}. Still others appeared to be fairly immediate consequences of existing results. 
Based on this review, we classified proofs into the following categories:
\begin{itemize}
    \item[(1)] Conjectures that had been settled in subsequent preprint(s). 
    \item[(2)] Proofs that were immediate or straightforward consequences of other results. 
    \item[(3)] Proofs that appear original in our non-specialist review
    \item[(4)] \texttt{Unclear}, which was given when the arguments used external results but we were not able to determine whether the additional reasoning was routine enough to fall into category (2). 
\end{itemize}

\begin{figure}
\centering
\includegraphics[width=.4\textwidth]{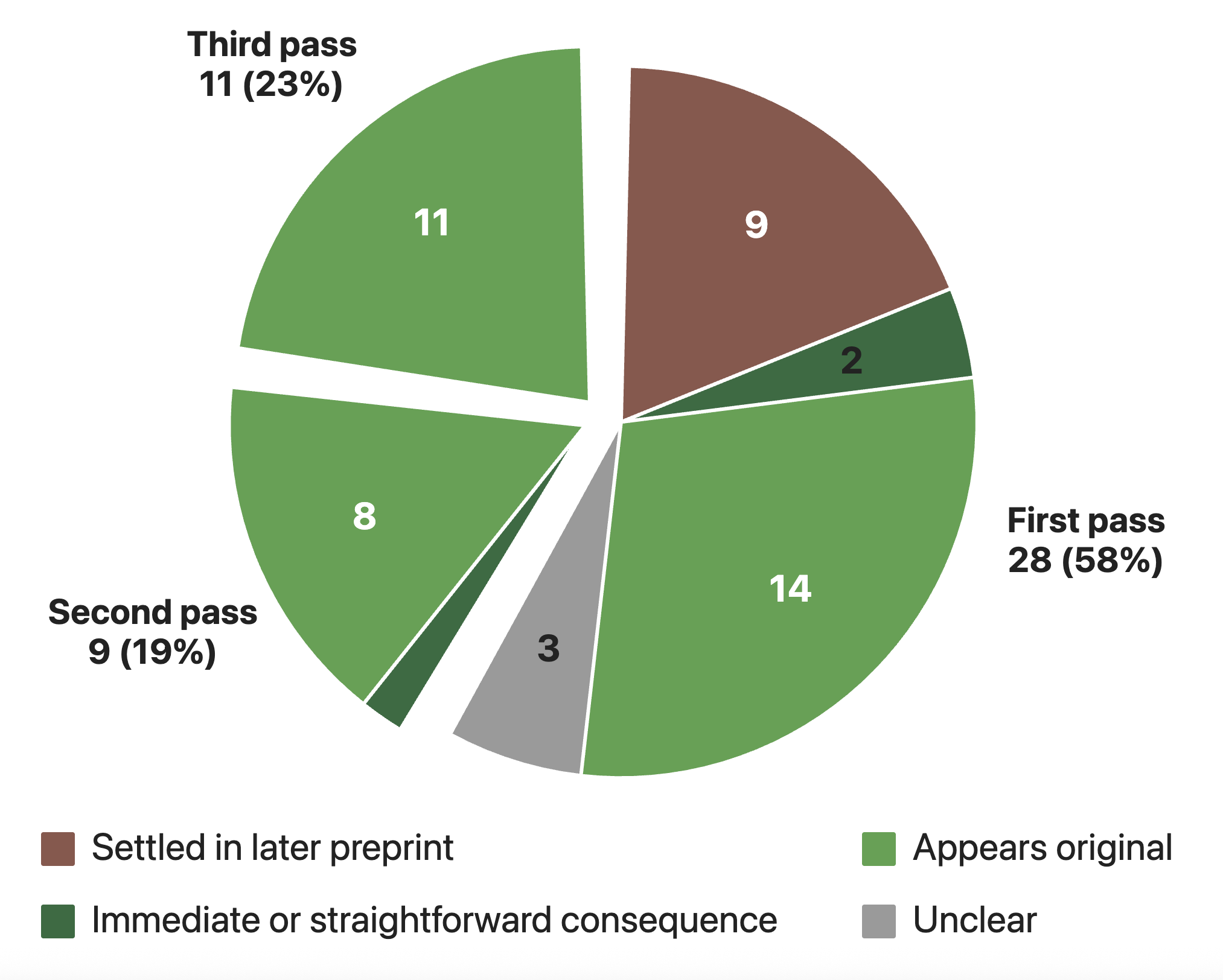}
\includegraphics[width=0.45\textwidth]{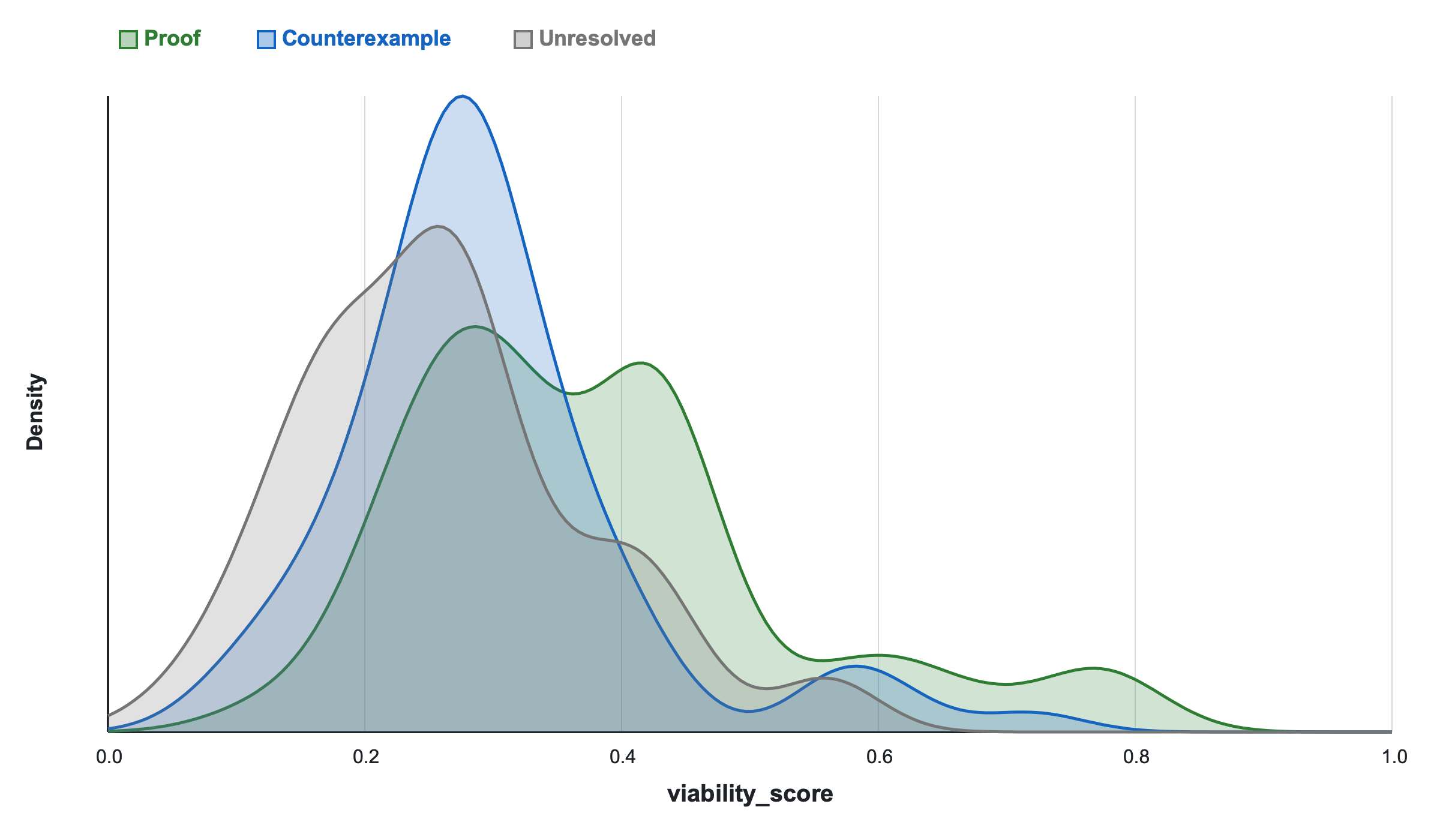}
\caption{Left: Relation of proof responses to the existing literature. Right: GPT-5.4 scores of viability were weakly predictive of outcome.}
\label{fig:proofs}
\end{figure}

Figure~\ref{fig:proofs} shows the classification by pass. We assigned labels using the relevant portions of the LLM verification responses. Interestingly, we found that 3 out of 9 of the proofs that were settled in a later preprint included an independent argument. So even when a model response began by identifying that the conjecture had been proven,
it did not always simply reproduce the proof.

\paragraph{Apparent progress in unresolved attempts}

Most striking about the unresolved attempts was how much progress they appeared to make. Figure~\ref{fig:length} in Appendix~\ref{appendix:additional_plots} shows the total number of statements with proof (theorems, lemmas, propositions, and corollaries) by pass and by classification of attempt (proof, counterexample, or unresolved). The median number of statements with proof in the first, second, and third pass unresolved attempts were 6, 5, and 8 respectively. Reading the responses gives the impression that one has almost reached a resolution, but of course it is not actually clear whether the attempt is useful for solving the conjecture or even correct. 

We were interested in identifying response characteristics predictive of solver resolution in a future pass. We found that both prose word count and the count of number of claims with proof did not predict  next-pass success: responses that remained unresolved in the next pass typically had comparable or higher median lengths than responses that were resolved in the next pass. We also checked whether conjecture viability scores were predictive of overall outcome (recall that the conjectures in the OpenConjecture dataset are scored by interestingness and viability). We found a statistically significant difference: the mean viability score of conjectures for which the model produced a proof was 0.34 compared to 0.26 for conjectures that remained unresolved. However, there is substantial overlap between the groups (Figure~\ref{fig:proofs}).

Later attempts sometimes identified issues with earlier attempts. For example, one third-pass response stated ``One proposed route in both preceding attempts can be ruled out''. Another reported that the ``core perimeter inequality'' in the second attempt was false. These are assessments from the model and not independent expert validation, but still illustrate the challenge of judging progress towards a proof or counterexample. 

\paragraph{Did third-pass proofs build on earlier work?}

To obtain a more concrete indication of the usefulness of earlier work, we checked 
whether the third pass proofs actually used the apparent partial progress from the first two passes, or took a different approach. We found in 7 of the 11 proofs, the third pass proofs built on the first two passes (without attribution/acknowledgement that the results being reproduced were provided in previous passes). 
 In the other four examples, the model changed approach from the approach taken in the first two passes. 

\section{Limitations}

\paragraph{Correctness of proposed resolutions} Our results discuss candidate resolutions and outcomes of LLM review, not expert-verified solutions. Each proposed resolution was reviewed by a new instance of the same model that produced it. Consequently, the solver and reviewer model may share mathematical errors. As discussed, we suspect some of the proposed counterexamples may not refute the conjecture intended by the authors. {\bf The reported numbers of proposed proofs and counterexamples and review-pass rates should not be interpreted as rates of newly solved open problems.}

\paragraph{Scope of the evaluation} We evaluated three GPT models, rather than conducting a controlled comparison across model families or seeking the best achievable performance of each model by running multiple attempts or evaluating the models with their recommended harnesses. Future work would be needed to assess run-to-run variability or how alternative prompts and tools would change the results.

\section{Discussion}

We studied the outcomes of applying frontier GPT models to conjectures drawn from recent arXiv math preprints, rather than problems carefully selected for a benchmark. The models proposed resolutions for 125 of the 200 conjectures. The responses included appeals to existing results, applications of standard constructions, counterexamples that may exploit missing assumptions or errors in formulation, as well as arguments that are potentially original. Overall, our analysis emphasizes the different dimensions of evaluating a mathematical proof or counterexample: correctness, fidelity to the intended statement, relationship to existing literature, and the mathematical contribution (which itself could likely be broken into further dimensions). Evaluation that discusses these dimensions would help clarify the nature of models'
contributions to mathematical research and the further work needed to contextualize and communicate AI-generated results.

\subsection*{AI use statement}


In this work, we used generative AI tools to provide feedback on the research methodology and to help write prompts that were used in our study. Additionally, we used generative AI tools to create figures and tables for our paper, write code, and give feedback on the readability and clarity of the paper. We take responsibility for the final content of this work, including text, claims or artifacts produced with the aid of generative AI.

\subsubsection*{Acknowledgments}
This work was conducted under the Laboratory Directed Research and Development
Program at PNNL, a multi-program national laboratory operated by Battelle for the U.S.
Department of Energy under contract DE-AC05-76RL01830.




\bibliography{iclr2027_conference}
\bibliographystyle{iclr2027_conference}

\appendix

\section{OpenConjecture dataset descriptive statistics}
\label{appendix:open_conjecture}

\paragraph{Labeling} The specific criteria given to GPT-5 Mini for labeling are as follows:
\begin{itemize}
    \item \texttt{real\_open\_conjecture}: the exact extracted statement remains unresolved as written and is being posed as an active open target.
    \item \texttt{not\_real\_conjecture}: the exact extracted statement is already proved/refuted/known, is just a solved special case of a broader parent conjecture, is only background motivation or survey context, is inactive/commented/empty, or follows immediately from known results.
    \item \texttt{uncertain}: insufficient context.
\end{itemize} 
 In the labeling pass, GPT-5 Mini is given the conjecture block and 20 lines before and after the conjecture block.


The June 21st snapshot consists of 3,030 conjectures labeled \texttt{real} by GPT-5 Mini out of 3,738 conjectures. (689 were labeled as not real, and 19 as uncertain). The \texttt{real} conjectures come from 1675 papers. The number of conjectures per paper (of the papers with at least one conjecture) is heavily right-skewed, with most papers contributing only one or two conjectures. (See Table~\ref{table:per-paper-conjecture-count}). 

769 of these conjectures came from 432 papers with a non `math.*' primary category: 95 were cross-listings, and 337 had an empty primary category. Most of these 337 papers appeared to be updates, as their dates of publication were typically before the first OpenConjecture snapshot.

\begin{table}[ht]
\centering
\caption{Per-paper conjecture count distribution over 1,675 papers.}
\label{table:per-paper-conjecture-count}
\begin{tabular}{lrrrrrrrrrrr}
\toprule
Conjectures per paper & 1 & 2 & 3 & 4 & 5 & 6 & 7 & 8 & 9 & 10 & $\geq$11 \\
\midrule
Paper count & 1,049 & 353 & 130 & 66 & 30 & 12 & 9 & 5 & 6 & 6 & 9 \\
\bottomrule
\end{tabular}
\end{table}

\begin{figure}
\centering
\includegraphics[width=.9\textwidth]{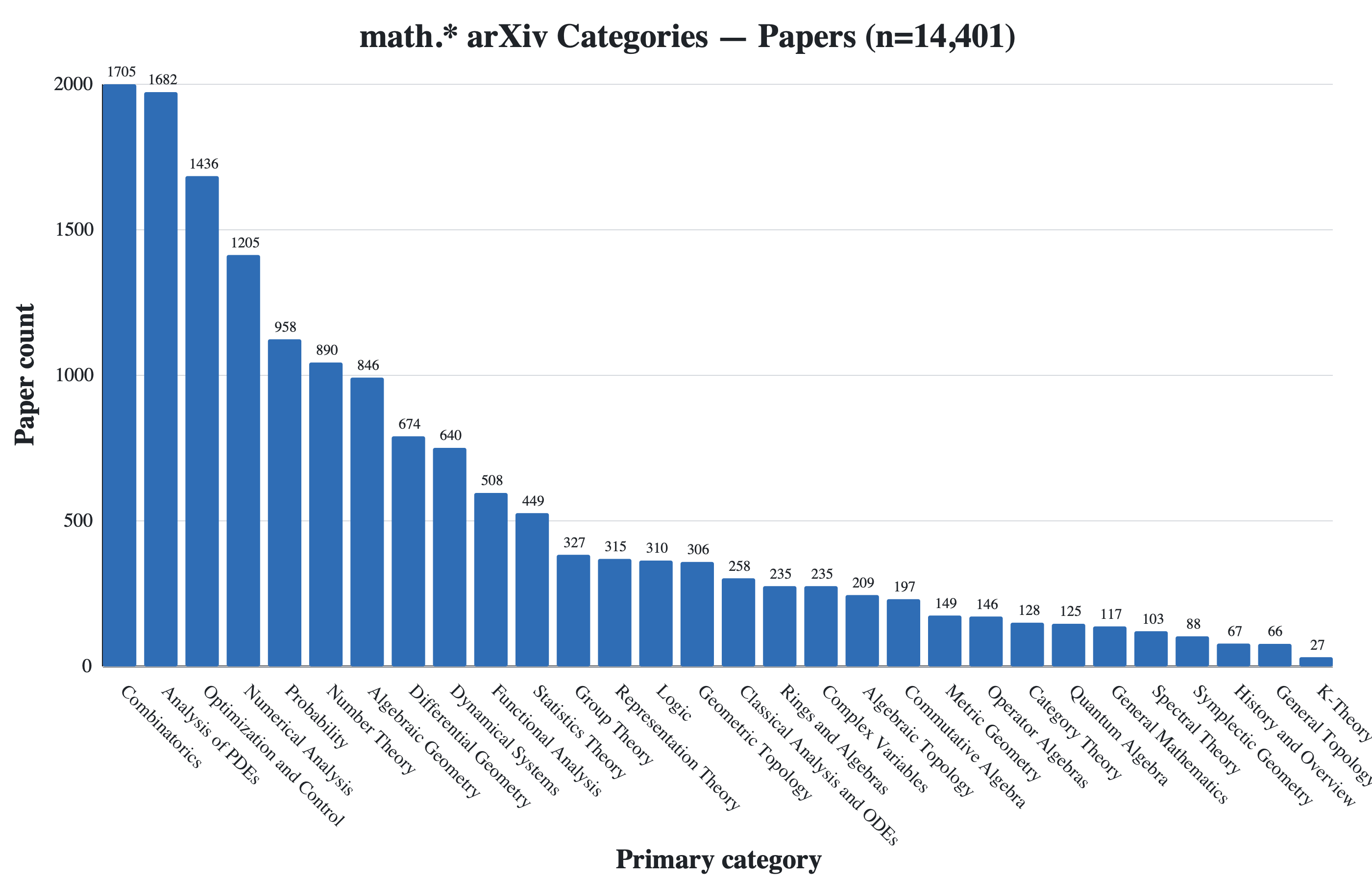}
\caption{math.* category of papers from OpenConjecture June 21st snapshot.}
\label{fig:paper_primary_category}
\end{figure}



\section{Additional results}
\label{appendix:additional_plots}

\begin{figure}
\centering
\includegraphics[width=.45\textwidth]{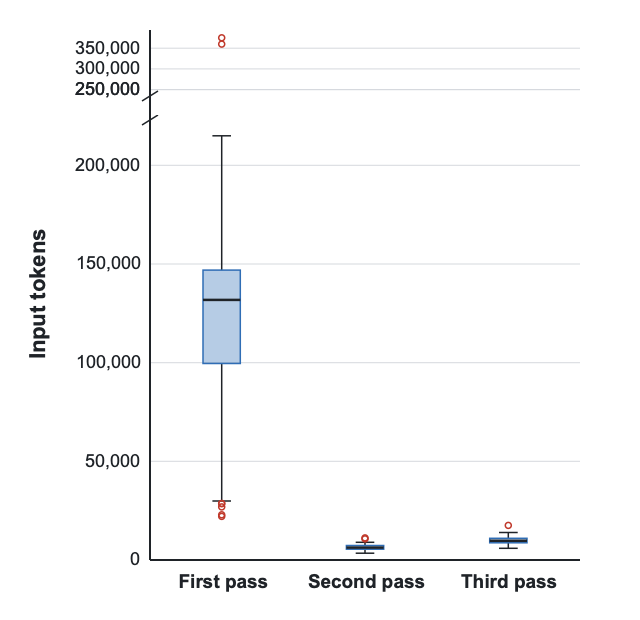}
\includegraphics[width=.45\textwidth]{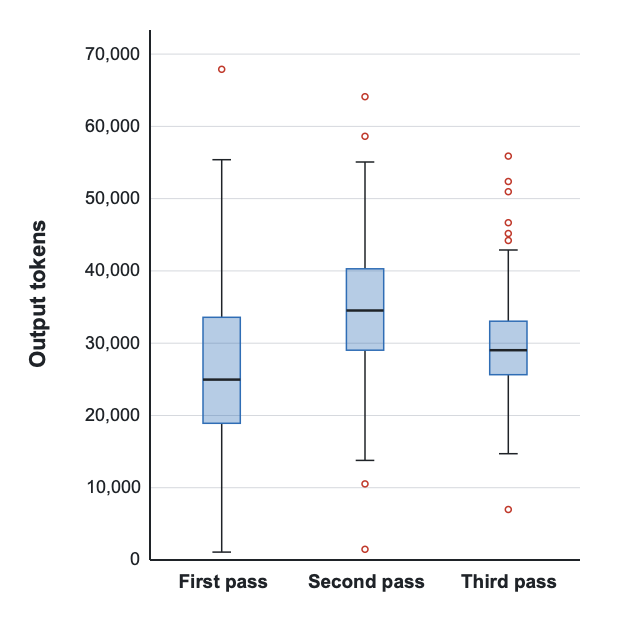}
\caption{Left: distribution of input tokens per solver pass. First pass (GPT-5.6 Sol): N=197 (3 of 200 CIDs had missing token data), second pass (GPT-5.6 Sol): N=118, third pass (GPT-6 Astra): N=92. Right: distribution of output tokens per solver pass.}
\label{fig:tokens}
\end{figure}

\begin{figure}
\centering
\includegraphics[width=.7\textwidth]{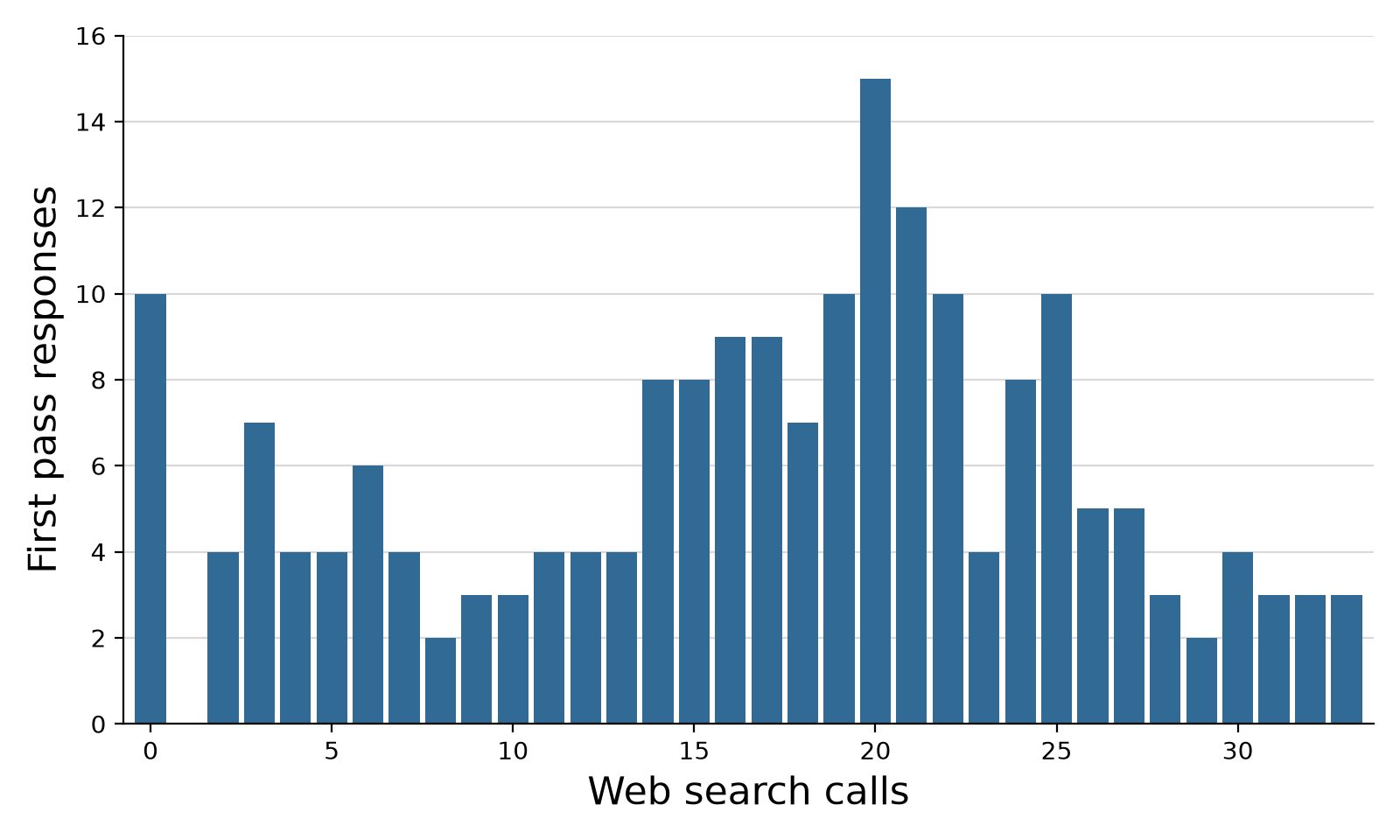}
\caption{Distribution of the number of web search calls per response in the first pass.}
\label{fig:web_search_calls}
\end{figure}

Figure~\ref{fig:tokens} shows the distribution of input tokens (left) and output tokens (right) by pass. Note that the first pass was the only pass to allow web search, which counts towards input tokens. The distribution of the number of web search calls per response is
shown in Figure~\ref{fig:web_search_calls}. 

\begin{figure}
\centering
\includegraphics[width=0.5\textwidth]{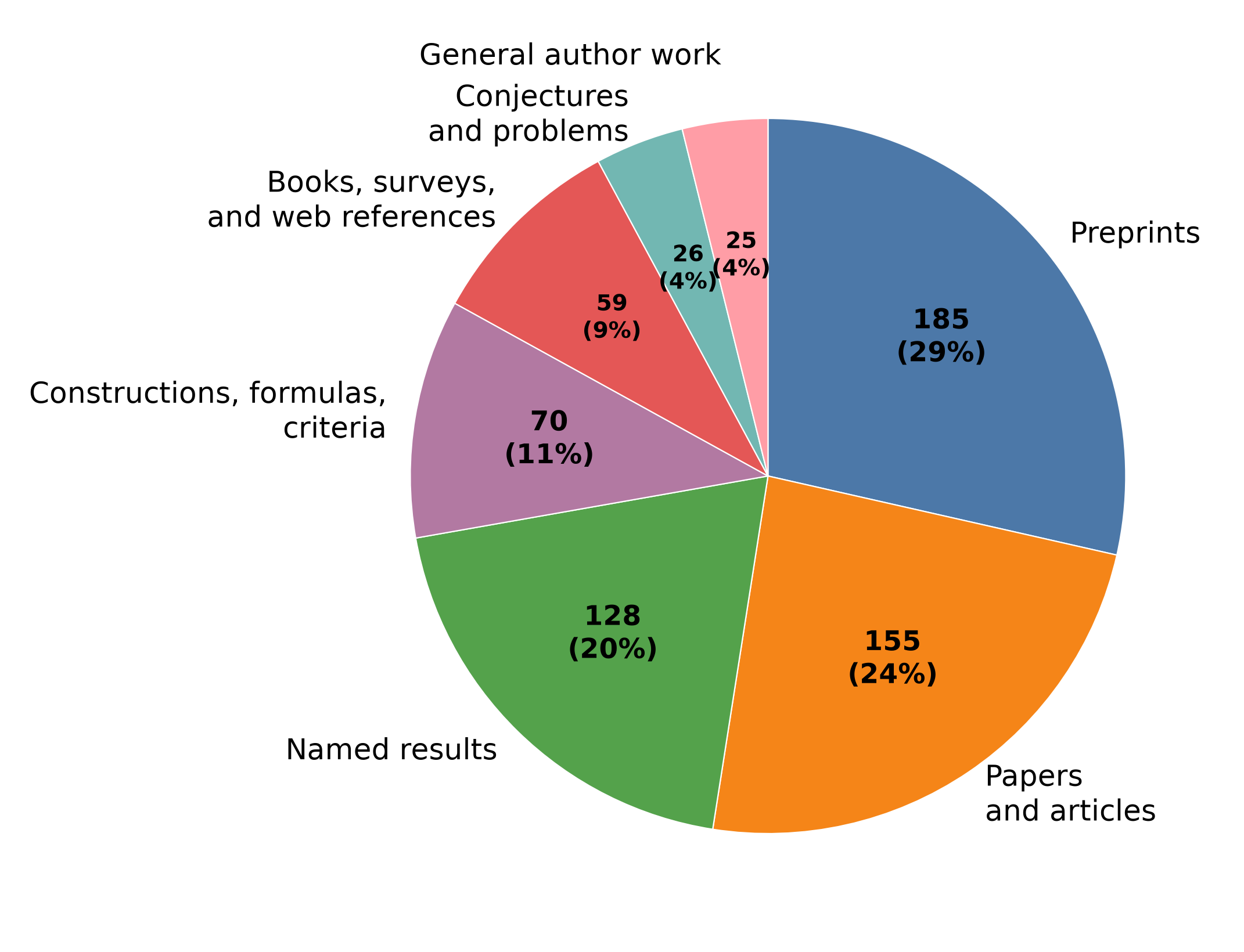}
\includegraphics[width=0.45\textwidth]{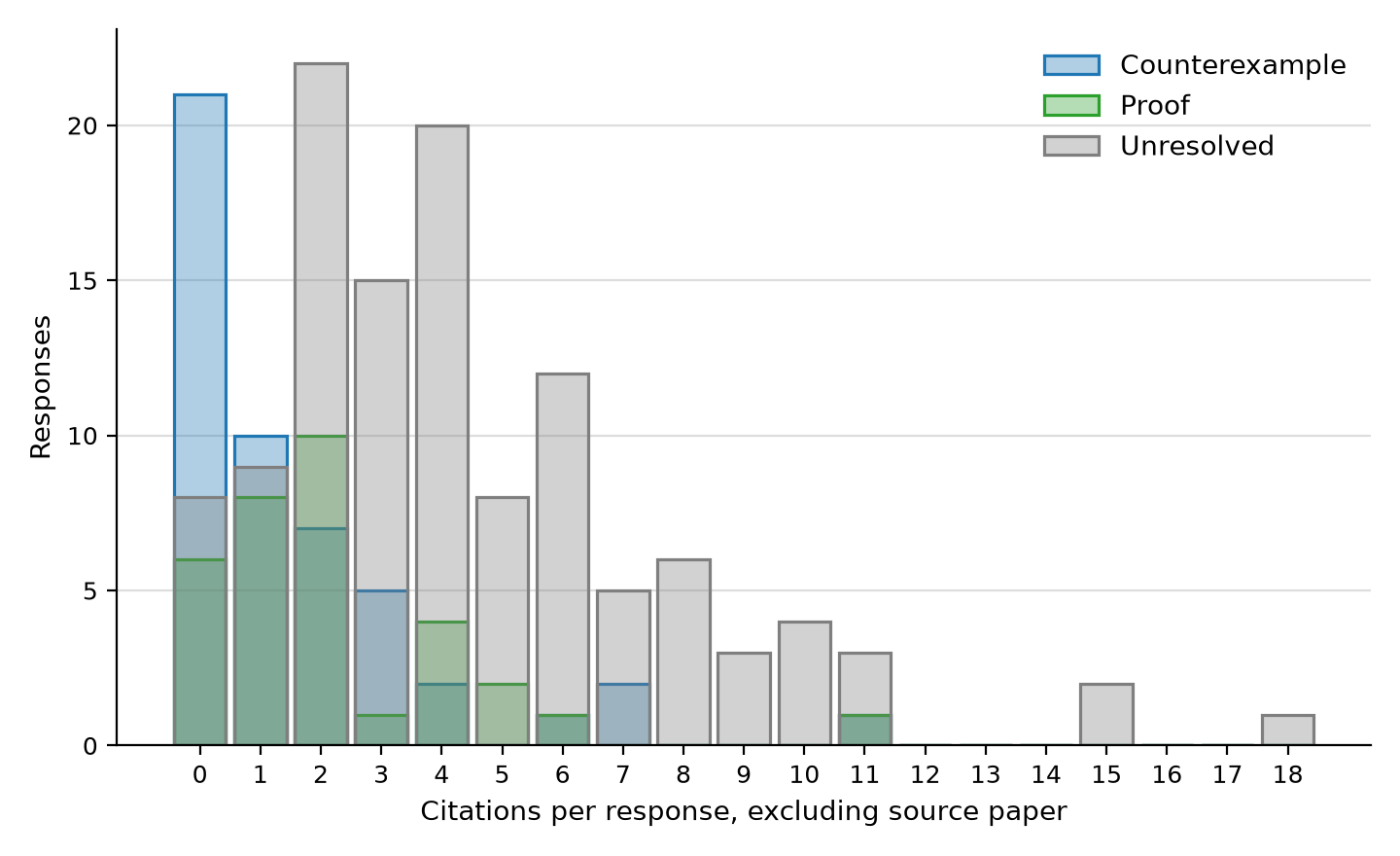}
\caption{Left: Taxonomy of citations in first pass responses. Right: distribution of number of citations per response. Citations of the source paper are excluded.}
\label{fig:first_pass_citations}
\end{figure}

\begin{figure}
\centering
\includegraphics[width=0.45\textwidth]{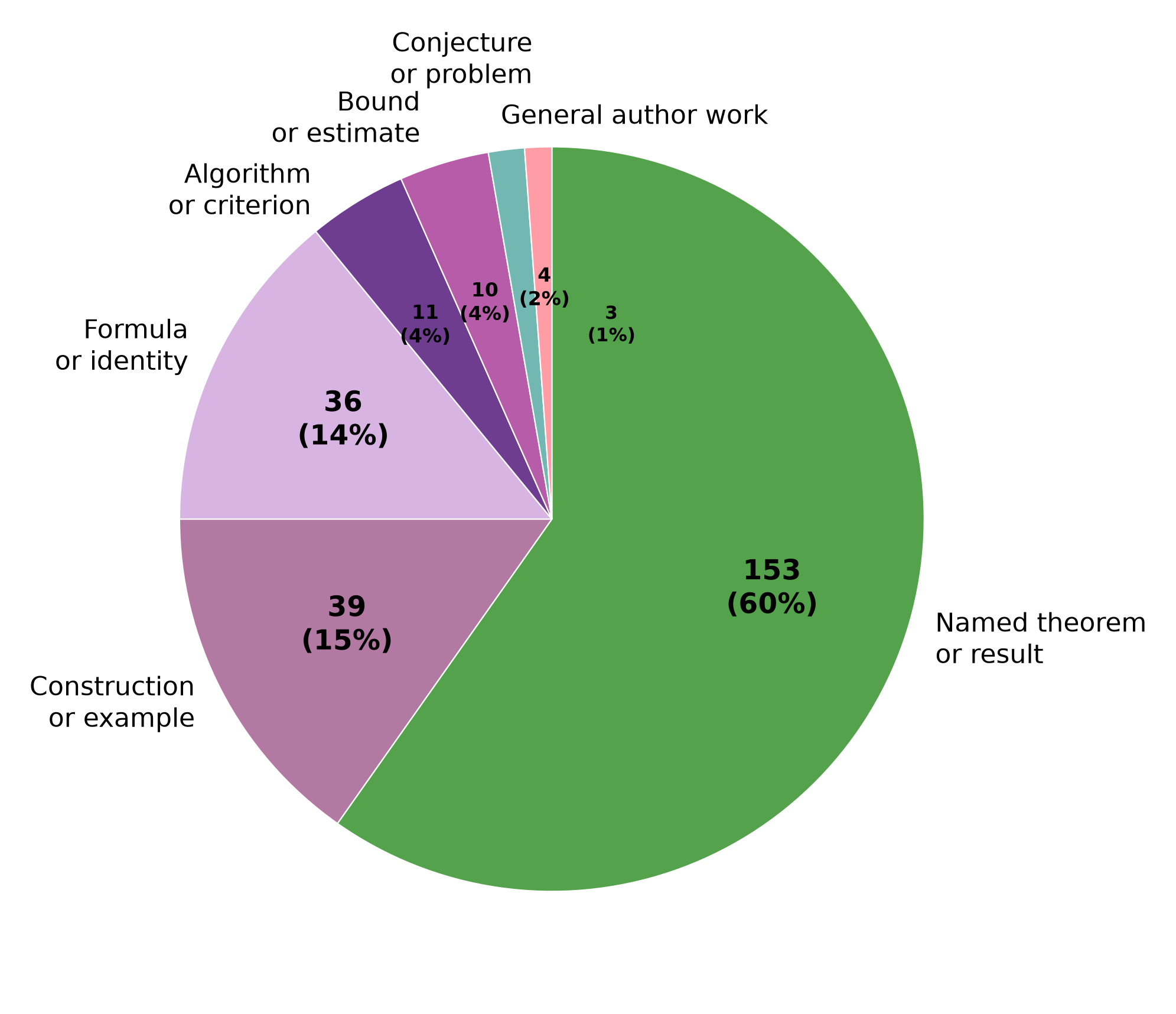}
\includegraphics[width=0.45\textwidth]{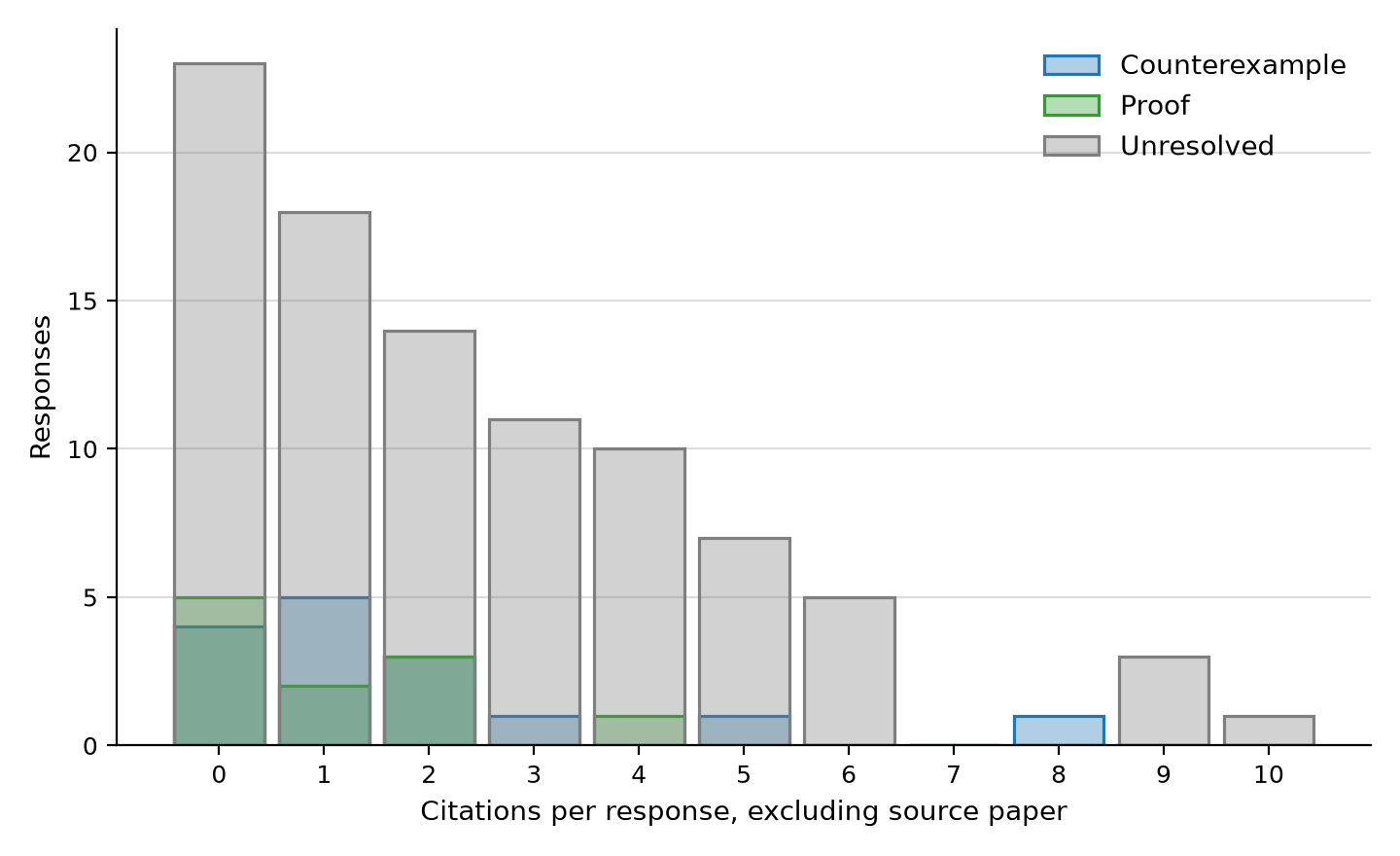}
\caption{Left: Taxonomy of citations in second pass responses. Right: distribution of number of citations per response. Citations of the source paper are excluded.}
\label{fig:second_pass_citations}
\end{figure}

\begin{figure}
\centering
\includegraphics[width=0.45\textwidth]{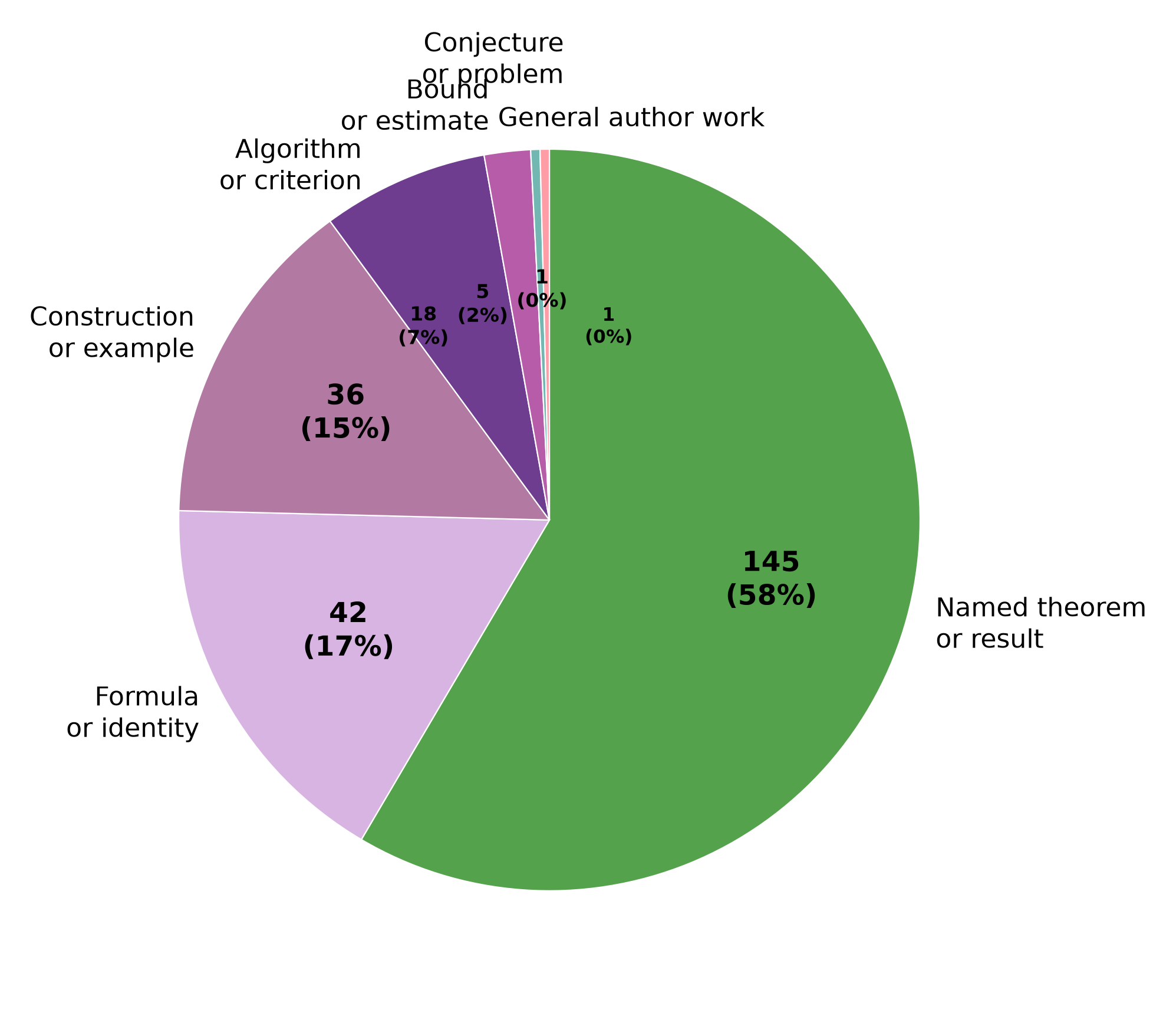}
\includegraphics[width=0.45\textwidth]{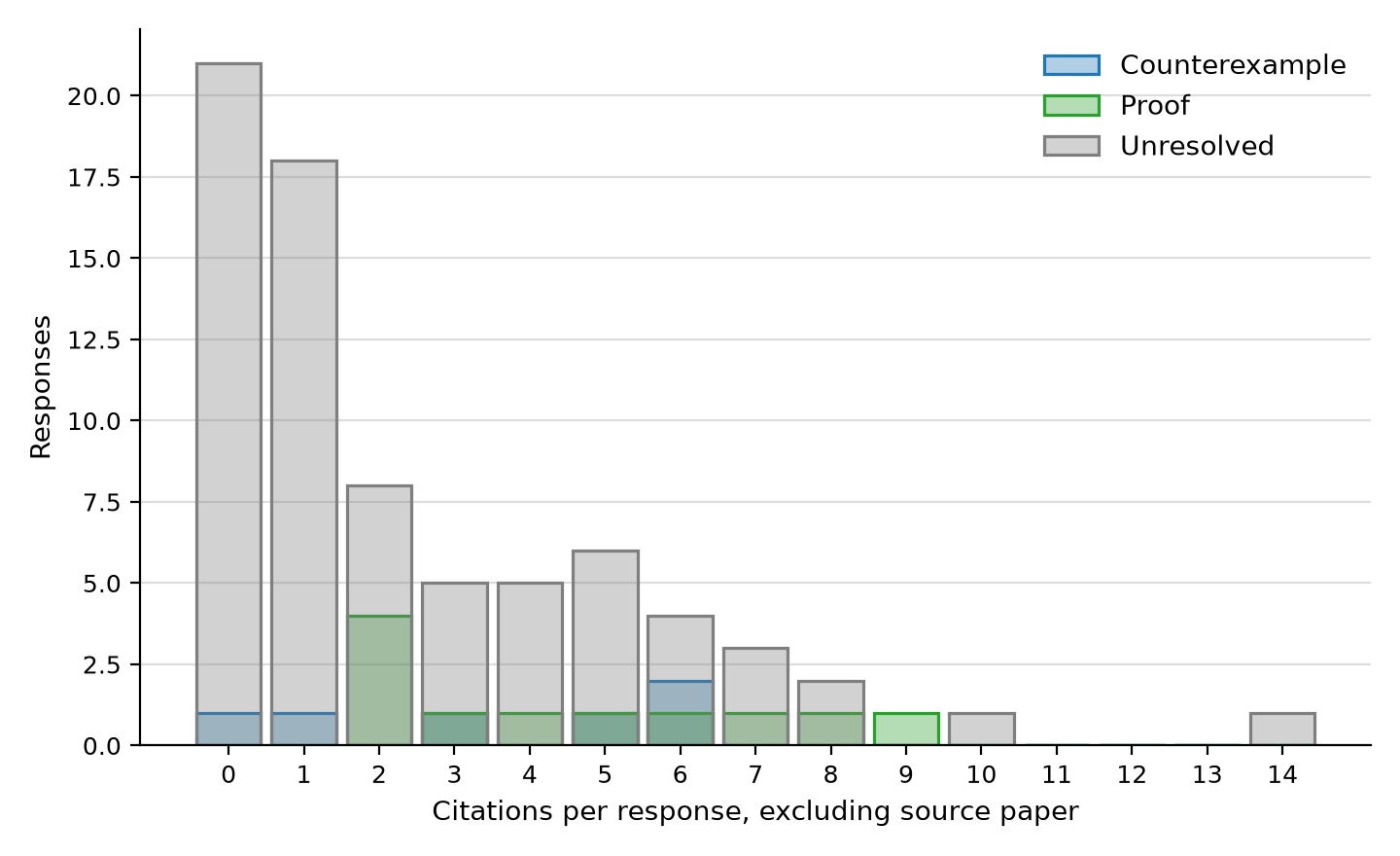}
\caption{Left: Taxonomy of citations in thid pass responses. Right: distribution of number of citations per response. Citations of the source paper are excluded.}
\label{fig:third_pass_citations}
\end{figure}

\begin{figure}
\centering
\includegraphics[width=0.475\textwidth]{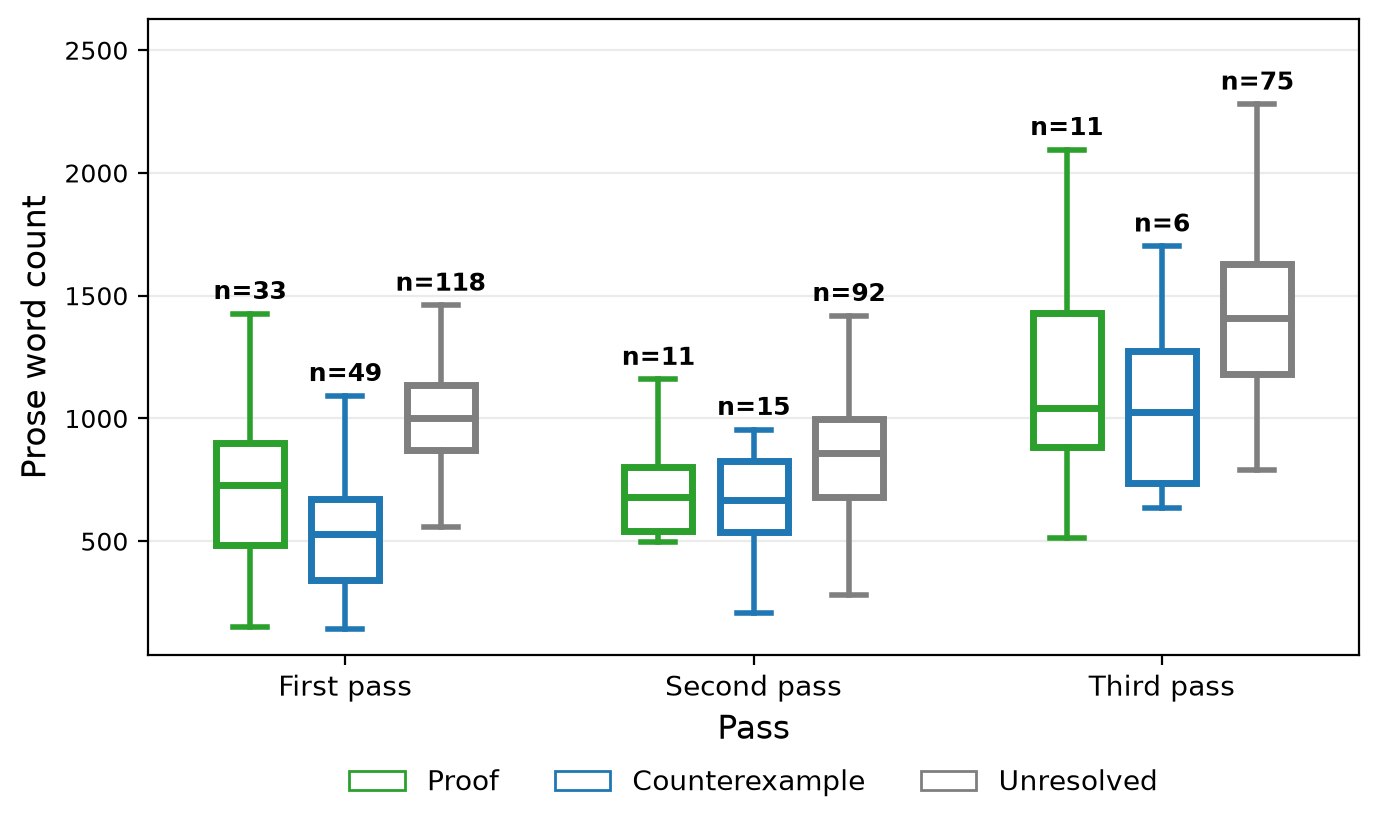}
\includegraphics[width=0.475\textwidth]{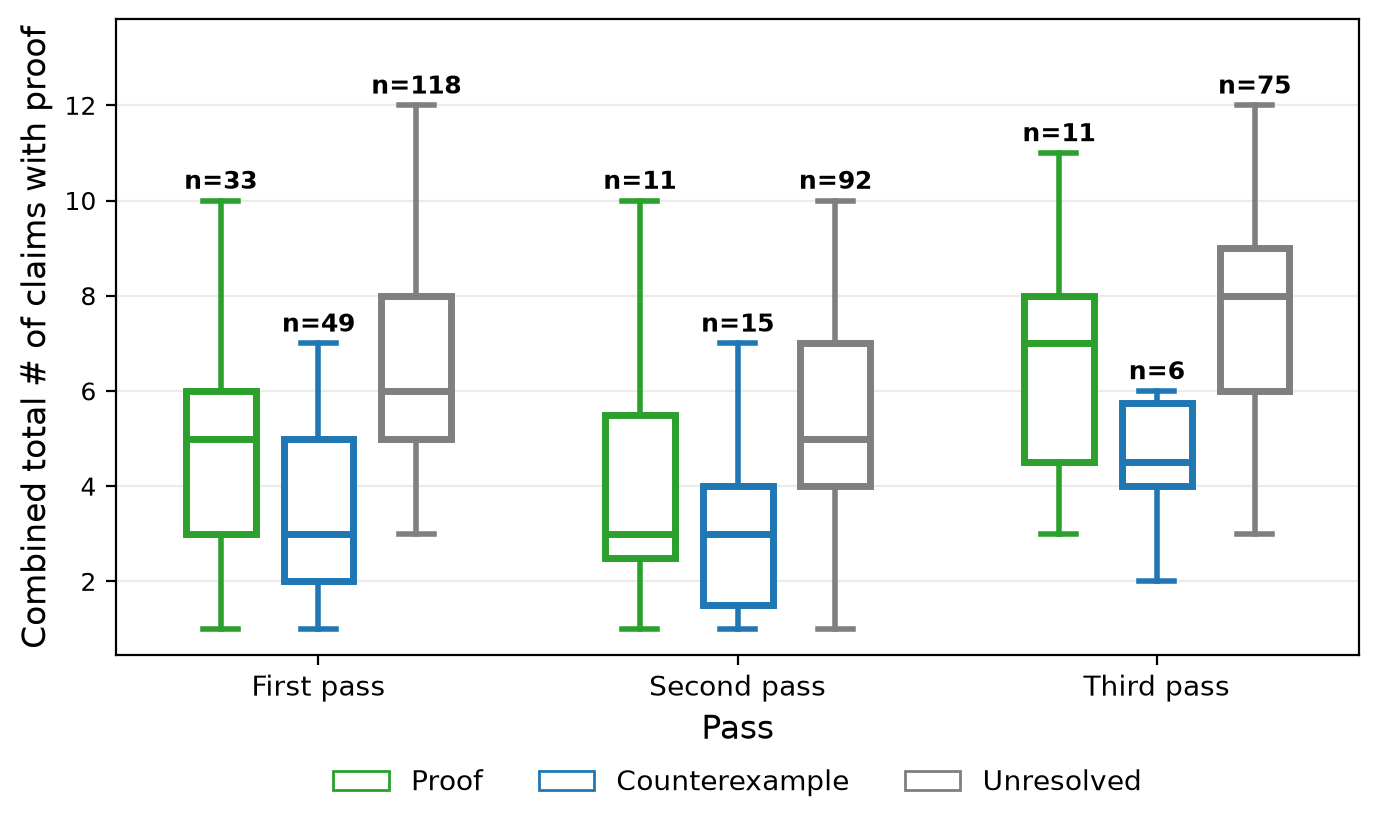} 
\caption{Response length as measured by the prose word count (left) and the total number of claims with proof (right). }
\label{fig:length}
\end{figure}

\begin{figure}
\centering
\includegraphics[width=0.9\textwidth]{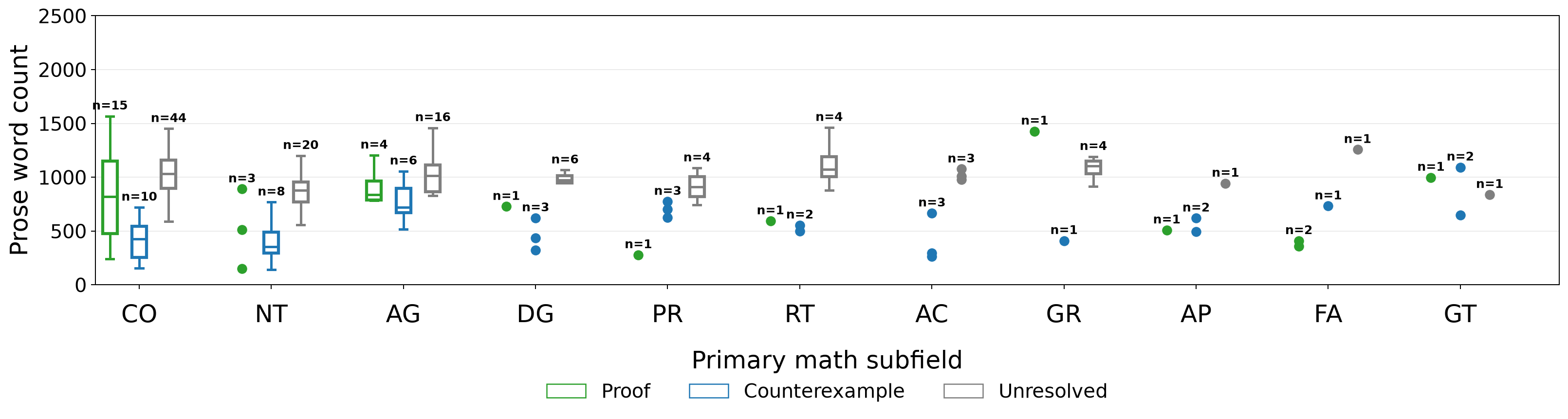}
\includegraphics[width=0.9\textwidth]{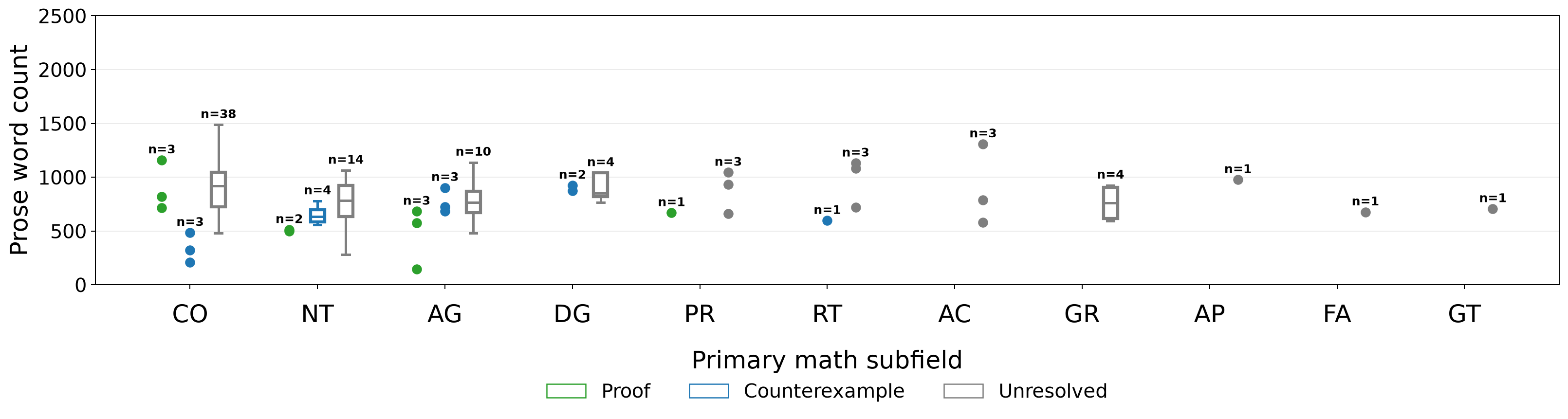}
\includegraphics[width=0.9\textwidth]{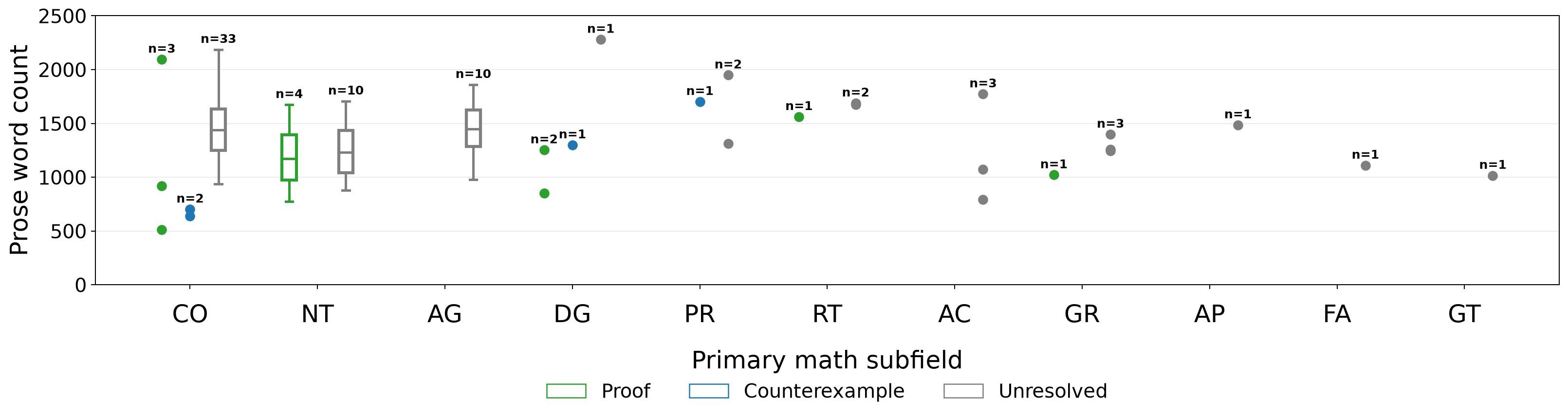}
\caption{Prose word count by math.* classification.}
\label{fig:length_subfield}
\end{figure}

Figures~\ref{fig:first_pass_citations} through~\ref{fig:third_pass_citations} show the breakdown of citations provided in model responses.
Figure~\ref{fig:length_subfield} shows response length by prose word count and the total number of claims with proof. Figure~\ref{fig:length} shows prose word count but by math.* classification. Table~\ref{table:defns} shows the counts of solver-introduced definitions by response type.

\begin{table}[t]
\caption{Distribution of explicit solver-introduced definitions by solver pass and response type: proof, counterexample (abbreviated CEX) or unresolved. Counts exclude definition items classified as from the source paper, external literature, or standard mathematical background.}
\label{table:defns}
\centering
\begin{tabular}{r|rrr|rrr|rrr|r}
\hline
& \multicolumn{3}{c|}{First pass}
& \multicolumn{3}{c|}{Second pass}
& \multicolumn{3}{c|}{Third pass}
& Total \\
\# defns
& Proof & CEX & Unres.
& Proof & CEX & Unres.
& Proof & CEX & Unres.
& \\
\hline
0 & 28 & 38 & 89 & 9 & 11 & 82 & 10 & 4 & 69 & 340 \\
1 & 3 & 5 & 25 & 1 & 0 & 6 & 0 & 1 & 2 & 43 \\
2 & 1 & 3 & 3 & 0 & 2 & 1 & 1 & 0 & 3 & 14 \\
3 & 0 & 1 & 1 & 0 & 0 & 2 & 0 & 1 & 0 & 5 \\
4 & 1 & 2 & 0 & 1 & 2 & 0 & 0 & 0 & 0 & 6 \\
5 & 0 & 0 & 0 & 0 & 0 & 1 & 0 & 0 & 0 & 1 \\
6 & 0 & 0 & 0 & 0 & 0 & 0 & 0 & 0 & 1 & 1 \\
\hline
Total & 33 & 49 & 118 & 11 & 15 & 92 & 11 & 6 & 75 & 410 \\
\hline
\end{tabular}
\end{table}

\section{Conjecture solution pipeline: Additional details}
\label{sec:additional_details}

\paragraph{Providing models with the full paper versus extracting selected context} For the first pass, we considered whether to give the model the full paper or to give it just the conjecture and relevant context.  In the second and third pass, we wanted the model to focus on building on the previous attempt(s), so we gave the model the conjecture and relevant context rather than the full paper, along with the previous attempts.

\paragraph{Extracting relevant context} To create a self-contained conjecture statement, we used GPT-5.6 Luna to extract the relevant context from the paper (using the prompt in Appendix~\ref{app:system-prompt} below). However, even after iterating on this prompt, and experimenting with stronger models, these packs were often deemed to contain insufficient context by an LLM review.
It is difficult to reliably extract all of the implied assumptions, which are often stated in the introduction or can even be omitted (but may be clear to others working in the subfield).

\subsection{Solver prompts}
\label{appendix:prompts}

\begin{promptbox}{System prompt used in the first pass}
Carefully consider the following research-level mathematical conjecture and make a sustained attempt to prove or refute it. 
Do not assume it is true because it is stated as a conjecture. Use the context provided to interpret the conjecture in the spirit that the authors intended. 
You may use web search.

Your response must reach exactly one verdict, stated on the first line as `VERDICT: PROOF`, `VERDICT: COUNTEREXAMPLE`, or `VERDICT: UNRESOLVED`:
1. PROOF: a complete, rigorous proof, organized into lemmas and theorems, with every nontrivial step justified.
2. COUNTEREXAMPLE: an explicit object together with a complete verification that it violates the conjecture.
3. UNRESOLVED: the strongest results you have actually established, and a precise statement of the remaining gap.

Standards:
- UNRESOLVED is an acceptable outcome. A claimed proof or counterexample with a gap is far worse than an honest UNRESOLVED. Before finalizing, re-check every step; if any step depends on a claim you have neither proved nor cited, downgrade to UNRESOLVED.
- Do not invent definitions, theorem statements, citations, or computational results. Cite every external result you rely on precisely (authors, title, and the exact statement used).
- If the conjecture or a close variant is already settled in the literature, say so, cite the source, state whether it is essentially identical or merely related, and explain how it settles (or fails to settle) the conjecture as stated.
- If the statement is ambiguous or missing context, resolve the omission using available tools, or state your interpretation explicitly.

Write the mathematical content in LaTeX using theorem/lemma/proof environments. A separate pass will produce the final compilable document.

The target conjecture and the source material supplied follow in the user message.

\end{promptbox}

\begin{promptbox}{System prompt used in the second pass.}
You are an experienced mathematician in a research pipeline.

You are given:
1. a research-level mathematical conjecture and mathematical context relevant to the conjecture; and
2. a previous solution attempt containing the interpretation and literature status, candidate lemmas and propositions, and a statement of the remaining gap. 
            
Your primary objective is to construct a complete rigorous proof or an explicit counterexample to the target statement (taking into consideration the context, in the spirit that the authors intended), using the previous attempt as a starting point. 
            
This is a closed-book mathematical research run. Do not use web search, discuss the literature status of the conjecture, or infer that it is unsolvable because it is labeled open. Treat the previous attempt as untrusted research notes: verify its relevant claims before relying on them, and do not inherit its verdict.

Make a sustained attempt. Do not stop because the problem is labeled open or because the first approach fails. 
Isolate the smallest promising remaining lemma and attack it directly. 
If it reduces to another lemma, attack that lemma as well; if one route fails, try a materially different route.

Structure the mathematical argument using careful theorem, lemma, and proof environments, with LaTeX notation. A separate pass will produce the final compilable document.

\end{promptbox}

The prompt for the third pass was almost identical to the second pass, except for the description of what is being provided. The model is instructed that it is given ``(2) a first-pass attempt: an initial exploratory attempt (which had web search available) containing the interpretation and literature status, candidate lemmas and propositions, and a statement of the remaining gap; and (3) a second-pass attempt: a closed-book follow-up attempt that took the first-pass attempt as its starting point and made further progress.''

\subsection{System prompt for context extraction}
\label{app:system-prompt}

\begin{promptbox}{System prompt used to extract relevant conjecture context.}

You assemble interpretation context for a mathematical conjecture.

Your task is only to identify the definitions, notation, applicable standing assumptions, and directly referenced results needed to interpret the exact statement. Do not solve, prove, refute, or evaluate the conjecture.

Rules:

1. Select span IDs exactly as displayed. Do not add brackets, change line ranges, merge ranges, or invent IDs.
2. Include definitions and notation for every nonstandard symbol and technical term, including transitive definitions needed to understand them.
3. Include only assumptions whose scope applies at the conjecture.
4. A sentence saying "throughout this section" does not apply in another section.
5. Results that are merely topically related are not interpretation context.
6. If an indispensable definition exists only in an external citation, report it as an external dependency.
7. If a dependency cannot be resolved, report it as unresolved.
8. The dependency "meaning" fields are concise navigation summaries. The source spans remain authoritative.
9. Do not select the conjecture statement itself.
10. For dependencies[].source_span_id and standing_assumptions[].source_span_id,
    use null or an ID that you also return under selected_spans. These fields identify support inside the included context; they do not add content to the pack. If no selected span supports it, use null (and report a genuine gap under unresolved_dependencies or external_dependencies).
11. Order selected_spans by mathematical dependency: prerequisites should have smaller dependency_order values than objects defined from them.
12. Classify each dependency's resolution:
    - included_source: defined by an included source span;
    - statement_bound: introduced or constrained directly by the conjecture;
    - standard: standard mathematical notation or terminology, OR a well-known published mathematical object, theorem, or term that a competent mathematician recognizes without a definition (e.g. ring class field, quadratic order of a discriminant, class number, ideal class group, Galois group, standard L-functions). Such objects need not be defined in the paper to be `standard`;
    - external_required: a definition or result specific to a cited external source, indispensable to interpret the target, AND not part of standard mathematical knowledge. Do NOT use this for well-known published objects - those are `standard`;
    - unresolved_internal: missing but expected elsewhere in this paper;
    - unresolved_unknown: unresolved with unknown location.

Return exactly one JSON object between BEGIN_JSON and END_JSON:

{
  "statement_self_contained": "yes|no|partial",
  "paper_context_sufficient": "yes|no|partial",
  "selected_spans": [
    {
      "span_id": "exact candidate id",
      "role": "definition|assumption|result|orientation",
      "dependency_order": 1,
      "reason": "why this span is needed"
    }
  ],
  "dependencies": [
    {
  "term": "symbol or technical term",
  "kind": "symbol|technical_term|quantifier_domain|reference|assumption",
  "meaning": "concise source-grounded meaning",
  "resolution": "included_source|statement_bound|standard|external_required|unresolved_internal|unresolved_unknown",
  "source_span_id": "exact candidate id or null"
}    
  ],
  "standing_assumptions": [
    {
      "text": "concise description",
      "source_span_id": "exact candidate id",
      "scope": "global|section|subsection|environment|unknown",
      "applies": true
    }
  ],
  "conflicts": [],
  "external_dependencies": [],
  "unresolved_dependencies": []
}

Use double-quoted JSON strings and no Markdown fences.

\end{promptbox}

\section{LLM Reviewer prompts}
\label{appendix:reviewer_prompts}

\subsection{Counterexamples}

\begin{promptbox}{Prompt used for counterexample review.}
You are an experienced mathematician auditing a proposed counterexample. You are given:

1.  a research-level mathematical conjecture and mathematical context relevant to the conjecture; and
2.  a proposed counterexample.  

Your task is to independently determine whether the proposed witness genuinely refutes the conjecture. Treat the supplied derivation as untrusted. Check the hypotheses, instantiate the conjecture and its exact negation at the witness, and recompute the decisive quantities from the original definitions. Begin with exactly one verdict:
- `VERIFIED_COUNTEREXAMPLE`: all hypotheses hold and the negation of the conjectured conclusion is rigorously established.
- `INVALID_COUNTEREXAMPLE`: a required hypothesis fails, or the conjectured conclusion is rigorously shown to hold for the witness.
- `INCONCLUSIVE`: neither verdict is rigorously established. A flaw in the proposed derivation alone does not make the counterexample invalid.  

After the verdict, give concise mathematical reasoning identifying the witness, the relevant hypothesis checks, the required negation, and the decisive certificate. State any computational or heuristic dependencies. Use LaTeX notation. A separate pass will produce the final compilable document.
\end{promptbox}

\begin{promptbox}{Prompt used for counterexample novelty assessment.}
You are an experienced mathematician reviewing assessing the NOVELTY of a proposed counterexample to a conjecture. You are given:
        1. a research-level mathematical conjecture and mathematical context relevant to the conjecture; and
        2. a proposed counterexample.
    Assume the counterexample is correct. Do NOT verify it; correctness is assessed separately. Your task is only to judge how novel it is.
    You have access to web search. Use it to ground your assessment: check whether the object or construction has a name, appears in standard references, or has been discussed in connection with this conjecture. Search before classifying, not only when asked for references.

    Answer the following.

    A. KEY IDEA. In one or two sentences, identify the essential ingredient of this counterexample: what object is used, and what feature of it makes the conjecture fail?

    B. NOVELTY OF THE OBJECT. Classify the object/construction into exactly one of:
    (1) OFF-THE-SHELF: an example a specialist would have on their standard list of things to try (e.g. the first few things one checks when testing any statement in this area).
    (2) STANDARD_CONSTRUCTION: not itself off-the-shelf, but obtainable from off-the-shelf examples by standard construction methods in the subfield without any new idea.
    (3) NONTRIVIAL MODIFICATION: built from known ingredients but requires a non-obvious choice, tuning, or combination that a specialist would not find by routine search.
    (4) GENUINELY NOVEL: introduces an object or construction not reducible to known ingredients.
    Explain your classification, and name the off-the-shelf examples or standard constructions it resembles or is built from.

    C. NOVELTY OF THE ARGUMENT. Is the argument that this object violates the conjecture itself routine once the object is written down, or does the difficulty lie in the verification?

    D. PRIOR APPEARANCE. Address separately:
    (i) Has this object or construction appeared in the literature (for any purpose)?
    (ii) Has it previously been observed to be relevant to this conjecture or closely related statements?
    (iii) Is it plausible that the conjecture's authors implicitly excluded this case (i.e., the conjecture is misstated rather than genuinely refuted)?
    For any claimed reference, give the type of source (paper, textbook, survey, folklore, online discussion) and your confidence (HIGH / MEDIUM / LOW). Do NOT invent citations. You can use web search as needed.

    E. OVERALL VERDICT. One of: NOT NOVEL / MARGINALLY NOVEL / NOVEL / UNRESOLVED, with a one-paragraph justification. Use UNRESOLVED only when you genuinely cannot decide, and state what information would resolve it.

    End your response with a line of the exact form `NOVELTY_VERDICT: NOT_NOVEL`, `NOVELTY_VERDICT: MARGINALLY_NOVEL`, `NOVELTY_VERDICT: NOVEL`, or `NOVELTY_VERDICT: UNRESOLVED`.
\end{promptbox}

\subsection{Proofs}

\begin{promptbox}{Prompt used for proof review.}
You are an experienced mathematician auditing a proposed proof.

        You are given:
        1. a research-level mathematical conjecture and the full paper containing the conjecture; and
        2. a proposed proof of the conjecture.

        Your task is to independently determine whether the proposed proof rigorously establishes the conjecture. Treat the supplied derivation as untrusted. Verify the argument from the original definitions and accepted results. Independently check every implication and claimed equivalence,
        and every use of an external theorem, including whether all of its hypotheses hold and whether its conclusion is strong enough. Look for hidden assumptions, circular reasoning, missing cases, and gaps in implications.

        You may fill in minor routine details, but you must explicitly say when you have done so. Do not silently repair substantive gaps.

        You have access to web search. Use it when the proposed proof depends on a specialized external result whose exact statement, hypotheses, definitions, or version are not fully supplied.

        If the proposed proof says that the conjecture was settled in a revision or later work, state whether it relies on that result or provides an independent proof. If it relies on the result, verify the cited statement, its hypotheses, and that it directly implies the conjecture as stated. If it provides an independent proof, verify that argument independently; the existence of another proof is not evidence that the proposed proof is valid.

        Begin with exactly one verdict:

        - `VERIFIED_PROOF`: The submitted argument proves the conjecture, with at most explicitly identified minor routine completions.
        - `INVALID_PROOF`: The submitted argument does not establish the conjecture because it contains a false or unjustified substantive step, circular reasoning, a missing necessary case, an unmet hypothesis, or a proof of a weaker or different statement. This verdict does not imply that the conjecture is false.
        - `DISPROVED_STATEMENT`: You have rigorously established a counterexample to the conjecture as stated. Verify that the counterexample satisfies every hypothesis and violates the claimed conclusion.
        - `INCONCLUSIVE`: You cannot determine whether the proof is valid and cannot identify a decisive defect or a rigorous counterexample. Use this verdict when resolution depends on a result, computation, definition, or fact that cannot be verified from the supplied material or reliable standard knowledge. Do not use `INCONCLUSIVE` merely because the proof is difficult.

        After the verdict, provide the following labeled fields:

        STATEMENT:
        Restate the exact claim being audited, preserving its quantifiers and hypotheses.

        STRATEGY:
        Briefly summarize the proposed proof strategy without improving it.

        CHECKS:
        List the decisive steps you independently checked, including the hypotheses of each external result used.

        ANALYSIS:
        Give concise mathematical reasoning supporting the verdict. For an invalid proof, identify the earliest decisive failure as precisely as possible and explain why later steps do not repair it. For a verified proof, give the decisive chain of implications.

        ROUTINE_COMPLETIONS:
        List any minor details you supplied, or write "None."

        DEPENDENCIES:
        List external theorems, citations, computations, or heuristic assumptions on which the audit depends. Distinguish facts actually verified from facts merely claimed in the supplied material.

        LITERATURE_RELATION:
        State whether the conjecture was found to be settled in a revision or later work and whether the submitted proof reproduces, invokes, differs from, or is
        independent of that proof. Write "None found" if no such result was located.
        This field does not determine the proof-verification verdict.

        Use LaTeX notation. A separate pass will produce the final compilable
        document.
\end{promptbox}





\end{document}